# Adaptively Weighted Top-N Recommendation for Organ Matching

Parshin Shojaee

Grado Department of Industrial and Systems Engineering, Virginia Tech, Blacksburg, VA 24061, USA, parshinshojaee@vt.edu

Xiaoyu Chen

Grado Department of Industrial and Systems Engineering, Virginia Tech, Blacksburg, VA 24061, USA, xiaoyuch@vt.edu

Ran Jin*

Grado Department of Industrial and Systems Engineering, Virginia Tech, Blacksburg, VA 24061, USA, jran5@vt.edu

Reducing the shortage of organ donations to meet the demands of patients on the waiting list has being a major challenge in organ transplantation. Because of the shortage, organ matching decision is the most critical decision to assign the limited viable organs to the most "suitable" patients. Currently, organ matching decisions were only made by matching scores calculated via scoring models, which are built by the first principles. However, these models may disagree with the actual post-transplantation matching performance (e.g., patient's post-transplant quality of life (QoL) or graft failure measurements). In this paper, we formulate the organ matching decision-making as a top-N recommendation problem and propose an Adaptively Weighted Top-N Recommendation (AWTR) method. AWTR improves performance of the current scoring models by using limited actual matching performance in historical data set as well as the collected covariates from organ donors and patients. AWTR sacrifices the overall recommendation accuracy by emphasizing the recommendation and ranking accuracy for top-N matched patients. The proposed method is validated in a simulation study, where KAS [60] is used to simulate the organ-patient recommendation response. The results show that our proposed method outperforms seven state-of-the-art top-N recommendation benchmark methods.

# 1  Introduction

According to the World Health Organization (WHO), "Transplantation is the transfer of human cells, tissues or organs from a donor to a recipient with the aim of restoring function(s) in the body" [58]. Organ transplantation is a widely adopted and effective therapy for patients with end-stage organ failures and fatal diseases. In the United States, the number of successful organ transplants for all safely transplantable organs, such as kidney, liver, heart, lung, intestine, and pancreata, has increased roughly from 28,940 in 2006 to 36,529 in 2018 [61]. However, the total number of patients on the waiting list of all transplantable organs has increased from 94,441 in 2006 to 113,759 in 2018 [61]. Since demand greatly outstripped supply, organ shortage has become a critical challenge in the organ transplantation network management. Among all transplantable organs, kidney is the most demanded one. Figure 1 presents that roughly 93,661 patients from the United States were on the waiting list for kidney transplant in 2018, while only 21,167 kidney transplantations were performed in the same year due to the limited donations [62]. Therefore, the shortage of kidney transplants by the end of 2018 results in approximately 77.4% of patients remaining on the waiting list for viable organs [62]. Since 2006, kidney matching has undergone many revisions. The latest kidney matching revision has been performed in 2014 [60], and kidney shortage is affected afterward as shown in Figure 1. There is an opportunity to further improve this matching decision. Improved matching decision leads us to better use of the limited available organs and to better serve the needs from patients.

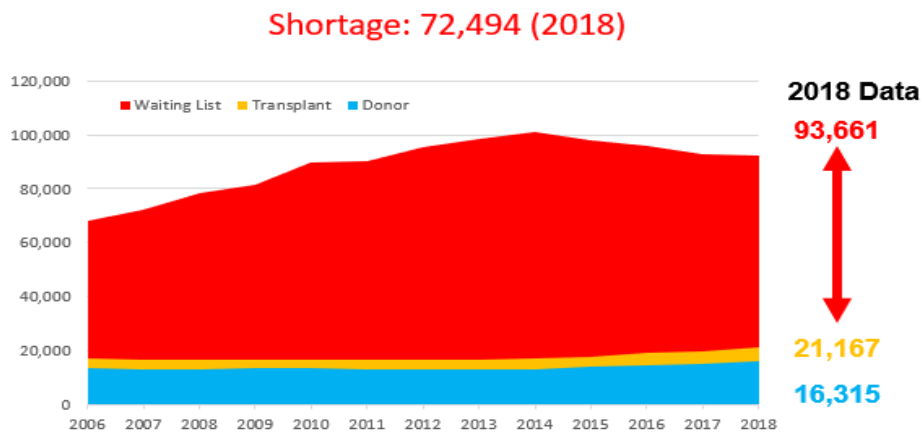

**Figure 1: Kidney shortage for transplantation from 2006 to 2018.**

The United Network for Organ Sharing (UNOS) is a non-profit organization that operates the Organ Procurement and Transplantation Network (OPTN) which is responsible for organ matching decisions in the United States. Taking kidney transplantation, donors are either living donors (e.g. family members or friends) [63] or deceased donors. The majority of patients cannot find any living donors; thus, they will join a pool of candidates waiting for kidneys from proper deceased donors. When an organ becomes available from a deceased donor, there are thousands of medically compatible and available patients on the waiting list to whom the organ can be allocated. This problem is called the *organ matching* problem. In current practice, a matching score [64] will be calculated to determine the assignment of an available organ to waiting patients. The matching score indicates the priority on the waiting list. Then, all medically compatible candidates on the waiting list are ranked according to their matching scores. First, an available organ will be offered to the candidate with the highest matching score. If this candidate rejects the offer, the organ will be offered to the candidate with the second highest matching score. This offering in orders will continue until the organ is accepted by the selected patient. As a result, the assignment of organs highly
2

depends on the top matching scores of the patients, instead of only the highest matching score (i.e., the best matched patient), or all matching scores of the patients on the entire waiting list.

In the current practice of OPTN, matching score is calculated from an algorithmic model based on the first principles in a deterministic manner [64]. This matching score model may not be in line with real post-transplant matching performance and cannot perform well in case high proportion of information from donors and patients is missing (i.e., missing covariates and response). Therefore, since the resulting score may not reflect the actual post-transplant matching performance (e.g., patient's post-transplant quality of life (QoL) [32] or graft failure measurements), there is a pressing need to improve the matching decision. An improvement opportunity lies in considering the real post-transplant matching performance historical data as well as specific covariates of both organs and patients. Making such a decision based on limited post-transplant matching performance data is not a trivial problem. On the one hand, there was typically only one or two patients who received the organ donated from one donor out of many potential patients in the historical database, thus there were limited data of post-transplant performance data recorded. On the other hand, when determining the organ assignment, it is more important to accurately predict the post-transplant performance for those patients with top scores (i.e., a few top-ranked patients), rather than the prediction of all patients on the waiting list. We refer to prediction accuracy of the top-ranked patients as top-N accuracy and top-N ranking accuracy. The top-N accuracy reflects the probability of including top-ranked patients in the recommendation. The top-N ranking accuracy reflects the ranking accuracy among the top-ranked recommended patients. Due to these challenges, one needs to propose new methods for organ matching decisions that can exploit the sparse real data from post-transplantation performance in the historical database as well as covariates from organs and patients to improve the matching decision.

In this paper, we formulate the organ matching decision-making as a top-N recommendation problem. In organ matching problem, a recommendation matrix is defined, with each row as a viable organ, each column as a patient on the waiting list, and each entry as the post-transplant response of each organ-patient assignment. An available organ will have the covariates (e.g., donor's age, BMI, blood type, quality of organ, cause of death, geographical location, etc.), and each patient on the waiting list will have covariates (e.g., patient's age, BMI, blood type, waiting time, urgency measure, history of hypertension, history of diabetes, geographical location, etc.). The collected covariates from donors and patients may differ based on the organ type (i.e., kidney, liver, lung, etc.). Recommender systems employ machine learning (ML) models (e.g., content-based, user-based, item-based, collaborative filtering, etc.) to deal with high percentage of missing entries [35]. However, the top-N accuracy and top-N ranking accuracy of existing recommender systems still need improvement in case the values of matching scores for top organ-patient assignments are close, or a high proportion of data is missing. In this paper, the top-N accuracy is measured by Hit Rate (HR) which is defined as the percentage of true top-N patients that are included in the top-N recommended patients. However, the top-N accuracy doesn't consider the rank of top-N patients, which is important for organ matching decision-making. Therefore, we also consider the top-N ranking accuracy by using Normalized Discounted Cumulative Gain (NDCG) which is defined as the accuracy of predicting the rank of each patient in the top-N list of recommended patients for each organ. We propose a recommender system called Adaptively Weighted Top-N Recommendation (AWTR) to improve both HR and NDCG of top-N recommended patients for an available organ.

AWTR formulates recommendations based on ($i$) implicit similarities among organs and patients (i.e., the low-rank property of post-transplant matching performance); and ($ii$) explicit similarities explained by covariates of organs and patients. In this model, the implicit similarities can be formulated by the low-rank estimation; and the explicit similarities can be formulated by using a sparse regression model based on the covariates [21]. The recommender system in the organ matching problem should improve the top-N accuracy (i.e., HR) and top-N ranking performance (i.e., NDCG), rather than the matching score prediction accuracy for all patients on the waiting list (e.g., Normalized Root Mean Square Error (NRMSE)). To achieve this goal, AWTR will introduce weighted penalties for each element of the organ-patient response matrix. These weights are considered (1) to decouple top-N patients corresponding to each organ from other patients, and improve top-N accuracy, and (2) to create discrepancy within top-N patients of each organ and enhance top-N ranking accuracy. AWTR will iteratively update these weights in order to highlight top-ranked patients corresponding to each organ via higher penalty in the loss function of its optimization problem. These weights will be updated by the sigmoid transformation of the latest predictions. The nonlinear sigmoid function has a high slope for smaller positive weights and a low slope for greater positive weights. This nonlinearity contributes to the decoupling of top-N patients from the rest patients with lower priority on the waiting list and restricts the weights to be between zero and one. The iterative adaptation of weights based on the recommendations of previous iteration will help the recommendation model to better identify top-N patients corresponding to each organ, and to predict their ranking more accurately (later validated in Figure 3). As a result, AWTR will improve the top-N recommendation's performance, hence, improving the organ matching decision-making.

The rest part of this paper is organized as follows. Section 2 provides literature review of existing organ matching scoring models for organ matching decision as well as the review of existing top-N recommendation models. In Section 3, we propose the AWTR model as well as the computation algorithm in order to estimate unknown parameters and make recommendations with high top-N HR and NDCG. Section 4 presents the simulation study to illustrate the performance of AWTR and its advantage over other top-N recommendation benchmarks in the state-of-the-art, in terms of widely adopted evaluation metrics for the top-N accuracy and top-N ranking accuracy. Finally, the conclusions and future work are summarized in Section 5.

## 2 Literature Review

The organ matching scoring model has attracted interests in the academic literature. First, Ruth et al. [38] conducted a simulation model to study the organ matching problem. Then, Righter [36], David and Yechiali [13] exploited the stochastic optimization assignment problem formulation to model the organ matching decision. Meanwhile, Zenios [57], Roth et al. [37], Segev et al. [42], and Ashlagi et al. [3] concentrated on different scoring models for kidney matching optimization problem. Later, Stegall [44] formulated the organ matching problem as an optimization problem with the objective function of maximizing organ utility and called it "The right kidney for the right recipient", while, Gundogar et al. [18] considered both maximizing utility and minimizing inequity objectives by proposing the fuzzy organ allocation score (FORAS). Then, Bertsimas et al. [4] used linear regression models to identify the weight of each criterion in the integrated multi-criteria matching score formulations. Tong et al. [48] developed a kidney matching model based on the patient's preference, and Ahmadvand and Pishvaee [2] studied a model-based Data Envelopment Analysis (DEA) for the kidney matching decision. All these aforementioned methods have contributed to the current organ matching decision-making process used in the United States by OPTN.

Recently, Stegall et al. [45] summarized all the variants of kidney matching systems in the United States from 2004 to 2014, which led to the current kidney allocation system (KAS) [60]. KAS is the latest significant revision of the kidney matching decision that happened in the United States by OPTN; and it's based on the integration of criteria from both donors and recipients [4]. These criteria depend on the organ type (e.g., kidney, liver, heart, lung, etc.), and will be defined for donors and patients based on the covariates collected from them that affect the particular organ matching. Later, Dongping et al. [15] identified more criteria from donors and patients with the Delphi method in three rounds to improve survival after kidney transplantation. Nosotti et al. [31] modified Delphi technique used in [15] to improve the identification of significant criteria in the kidney matching score. Taking kidney as an example, the current practice of OPTN is considering Estimated Post Transplant Survival (EPTS) [65], Kidney Donor Profile Index (KDPI) [66], Calculated Panel Reactive Antibodies (CPRA) [59], waiting time, medical emergency, and geographical distance as effective criteria that can be integrated into the multi-criteria calculation of kidney matching score. Both simulation/optimization-based and criteria-based matching score models may not reflect the real post-transplant matching performance and suffer from missing information in the data collected from both patients and donors. Due to the aforementioned limitations and small sample size of the post-transplant matching performance, a top-N recommender model based on the real post-transplant data is expected to improve the matching performance.

In general, the purpose of any recommender system [1] is suggesting the most suitable products or items to a specific user by analyzing historical data from prior ratings. In the recommendation problem, the ratings for different items are structured into a sparse response matrix with each row as a user, each column as an item, and each entry as the rating (i.e., recommendation score) of each user-item suggestion. Recommender systems are either formulated as a rating prediction or a top-N recommendation problem. Collaborative filtering (CF) is one of the most well-known recommendation models in which similar ratings come from similar user behaviors and similar items ( i.e., the low-rank structure) [46]. The CF algorithm usually returns the list of suggested items for each user by considering the highest overall accuracy in terms of rating prediction [14,23,24,27,39,40]. These CF models ignore the order of items in the corresponding top-N recommendation list. Over the past decade, various approaches have been proposed to emphasize improving top-N recommendations and their ranking quality [17,35,43]. These approaches are divided into four categories: neighborhood-based CF, model-based CF, ranking-based, and sparse linear methods.

In the neighborhood-based CF models, one should find similarities that exist among users/items [14]. For example, the item-based k-nearest-neighbor (ItemKNN) determines the similar items among already rated items, and recommends top-N items based on those similarities. However, since ItemKNN employs just a few local item characteristics, it is still limited in identifying the ranking of elements in the top-N list.

Model-based CF methods build recommendations based on specifically generated models. Simple matrix factorization (MF) [28] and pure singular-value-decomposition (PureSVD) [11] are two examples of model-based CF. The simple MF exploits the user-item similarities, i.e., low-rank structure, to extract the purchase patterns for each user-item entry, while the PureSVD represents users and items by the most suitable singular vectors of the user-item response matrix.

The ranking-based models rely on the ranking/retrieval criteria, where the top-N recommendation is treated as a ranking problem. Bayesian personalized ranking (BPR) [33] is an example of methods in this category that consider the maximum posterior estimator from the Bayesian analysis. The BPR estimator is utilized to measure all the differences between the rankings of rated items and the remaining items.

Based on the literature, the sparse linear methods outperform all the aforementioned three categories in terms of top-N accuracy and top-N ranking accuracy. Recently, a top-N recommendation method, called sparse linear method (SLIM) [30] has been proposed. SLIM learns a sparse aggregation coefficient matrix from the user-item response matrix by considering each item as a linear combination of all other items and adding L1 and L2 norms in the regularized objective function. SLIM has obtained better top-N accuracy compared to the other state-of-the-art methods. However, SLIM can only capture relations between items that are co-rated by at least one user. In addition, one of the intrinsic characteristics of recommender systems is high sparsity due to the small proportion of available historical ratings. As a consequence, the low-rank sparse linear method (LorSLIM) [10] has been proposed to overcome the above limitation by imposing a low-rank constraint into the objective function. LorSLIM outperforms SLIM in terms of top-N accuracy and top-N ranking accuracy when the user-item response matrix is highly sparse.

## 3 Adaptively Weighted Top-N Recommender System

AWTR is the proposed top-N recommendation model that can be employed in the organ matching decision because of concentrating on improving top-N accuracy and top-N ranking accuracy. The major assumption made in AWTR is that similar post-transplant matching performances come from similar organs or similar patients (i.e., the low-rank structure of the organ-patient response matrix), which will be validated in Section 4.

We firstly define a sparse organ-patient response matrix $Y \in \mathbb{R}^{m \times n}$, where $m$ is the number of available organs and $n$ is the number of patients on the waiting list. Each entry of this matrix is the post-transplant matching performance score corresponding to each organ-patient assignment. We assume that $Y$ is sparse, i.e., the majority of the entries are unknown due to the limited historical post-transplant matching performance data. The purpose of this recommendation system is to predict these missing entries with satisfactory accuracy. Based on the literature, PRIME [9] is a recommendation model that outperforms matrix completion [55] by incorporating information contained in the covariates $X \in \mathbb{R}^{mn \times p}$ as well as combining the implicit and explicit similarities. By adopting PRIME model structure, the recommendation score (i.e., the post-transplant matching performance) can be decomposed into additive effects of the low-rank matrix and the linear regression term. This model framework can be formulated as follows.

$$Y = R + \mathcal{A}(X\boldsymbol{\beta}) + E, \tag{1}$$

where $R \in \mathbb{R}^{m \times n}$ is the low-rank matrix that represents implicit similarities among organs or patients; the column vector $\boldsymbol{\beta} \in \mathbb{R}^{p \times 1}$ reflects coefficients of covariates from organs and patients; $E \in \mathbb{R}^{m \times n}$ is also the recommendation error matrix. The reshaping operator, denoted by $\mathcal{A}(\cdot)$, maps any vector from $\mathbb{R}^{mn \times 1}$ to the matrix in $\mathbb{R}^{m \times n}$ to ensure the consistency of dimensions among $X\boldsymbol{\beta}$, $R$, $Y$, and $E$.

As with PRIME [9], our proposed method takes the sparse organ-patient response matrix ($Y$) and the covariate matrix ($X$) as input, where $X \in \mathbb{R}^{mn \times p}$ and $p$ is the number of features. In other words, each row in the covariate matrix ($X$) is connected to each entry of the sparse organ-patient response matrix ($Y$). In this paper, the objective function (2) is proposed as an unconstrained optimization problem for the AWTR model.

$$\min_{W, \boldsymbol{\beta}, R} \frac{1}{2} \left\| \left( W^{\frac{1}{2}} \odot (Y - \mathcal{A}(X\boldsymbol{\beta})) - R \right) \right\|_F^2 + \lambda_1 \|R\|_* + \lambda_2 \|\boldsymbol{\beta}\|_1. \tag{2}$$

The proposed objective function (2) aims at estimating $W \in [0,1]^{m \times n}, \boldsymbol{\beta} \in \mathbb{R}^{p \times 1},$ and $R \in \mathbb{R}^{m \times n}$. The low-rank estimation of the organ-patient response matrix, denoted by $\widehat{R}$, represents the implicit similarities among organs and patients. To enforce the low-rank property of $R$, we regularize it with a nuclear norm (i.e., $\|R\|_*$) in the objective function, where $\|R\|_*$ is defined by the sum of its singular values as the continuous and relaxed approximation of a matrix rank [7]. Therefore, minimizing $\|R\|_*$ leads to a low-rank matrix $\widehat{R}$. The estimated coefficient vector, represented by $\widehat{\boldsymbol{\beta}}$, considers the level of relationships between the covariates and the post-transplant matching performance. To enforce the sparsity of $\boldsymbol{\beta}$, we add L-1 norm of $\boldsymbol{\beta}$ (i.e., $\|\boldsymbol{\beta}\|_1$) into the loss function. This comes from Lasso regression [47] which identifies significant features by incorporating the L-1 norm into the least squares loss function. Novelty of the proposed method lies in the estimated weight matrix, denoted by $W$. The weight matrix assigns a weight to each entry of the organ-patient response matrix and will be iteratively adapted by the sigmoid transformation of the previous iteration's recommendation [51]. The weight matrix enables us to emphasize the top-ranked patients corresponding to each organ with a greater weight penalty in the objective function. The Frobenius norm, denoted by $\|.\|_F$, represents the square root of the sum of squared errors (i.e., the difference between predicted and actual scores). So, minimizing $\|.\|_F$ reflects reducing the bias term or underfitting in the proposed model. The element-wise multiplication operator, represented by $\odot$, is used for the multiplication of two matrices $W^{1/2}$ and $Y - \mathcal{A}(X\boldsymbol{\beta})$ with the same dimensions.

In this objective function, $\lambda_1$ and $\lambda_2$ are tuning parameters that control the amount of shrinkage in the rank of the low-rank matrix, $\|R\|_*$, and sparsity of the coefficient vector, $\|\boldsymbol{\beta}\|_1$, respectively. If $\lambda_1$ takes small value, $\|R\|_*$ is allowed to take greater values, thus the number of independent columns will increase, i.e., the similarity among columns will decrease. The lower the similarity among rows or columns, the worse the recommendation accuracy. Also, if $\lambda_2$ takes small value, more features are allowed to be significant. As a result, we need to find the best balance of these tuning parameters, $\lambda_1$ and $\lambda_2$, by applying a two-level grid search in the 5-fold cross validation. This will help us to reduce the gap between training and testing errors and avoid overfitting.

Figure 2 illustrates a simple graphical representation of our proposed method. Based on this figure, AWTR takes the covariate matrix ($X$) and the sparse organ-patient response matrix ($Y$) as the inputs. In the organ matching decision-making problem, the covariate matrix ($X$) is generated by all the effective features from donors and patients and all the pairwise interaction of these features; and the sparse organ-patient response matrix ($Y$) contains the existing post-transplant matching performances for the organ-patient assignments. AWTR formulates the optimization problem (2) to estimate the required model parameters. To solve the unconstrained optimization problem (2), we will use the alternating direction method of multipliers (ADMM). In the solving process, the weight matrix will be adapted iteratively by the sigmoid transformation of the latest predictions. Finally, based on the estimated parameters from solving (2), a predicted organ-patient response matrix will be obtained to find the top-N ranked recommendations by sorting columns for each row.

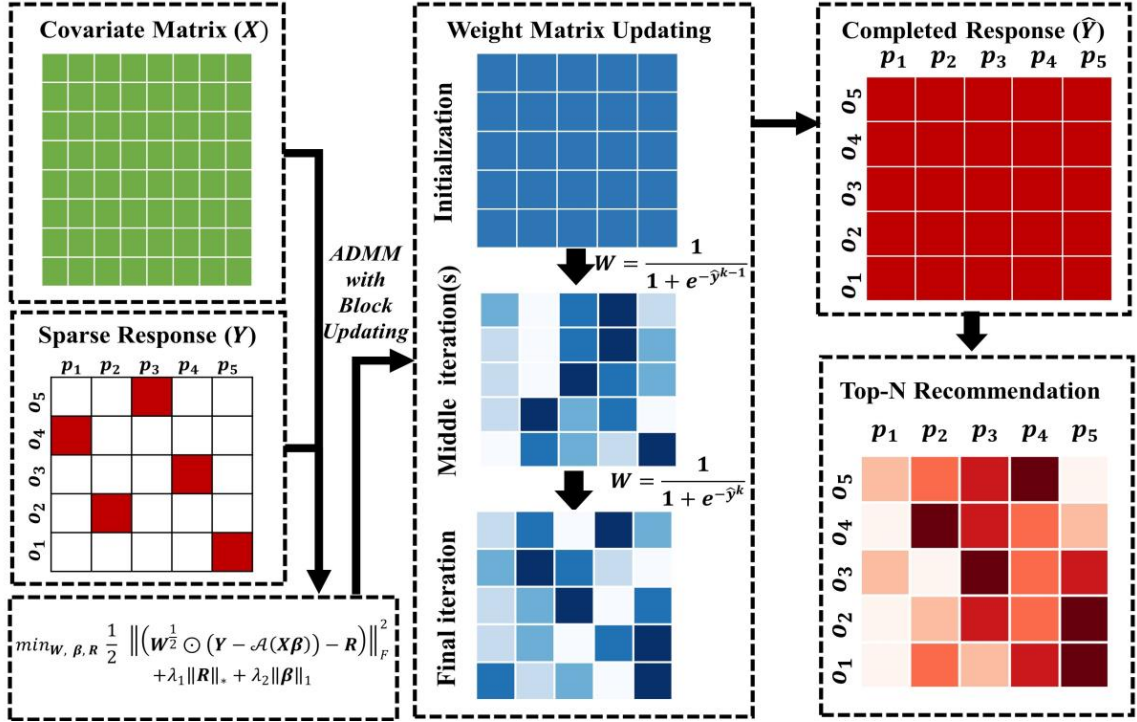

**Figure 2: A graphical representation of AWTR. Gradient colors in each row represent the ranking of recommendations.**

Motivated by the Split Bregman method [56], the augmented Lagrangian function of the objective function (2) can be written as follows.

$$\mathcal{L}(W, R, \beta, H, g, U, v) = \frac{1}{2} \left\| \left( W^{\frac{1}{2}} \odot \left( \mathcal{P}_\Omega(Y) - \mathcal{P}_\Omega(\mathcal{A}(X\beta)) \right) - \mathcal{P}_\Omega(R) \right) \right\|_F^2$$
$$+ \lambda_1 \|H\|_* + \lambda_2 \|g\|_1 + \langle U, R - H \rangle + \langle v, \beta - g \rangle \quad (3)$$
$$+ \frac{\mu_1}{2} \|R - H\|_F^2 + \frac{\mu_2}{2} \|\beta - g\|_2^2 ,$$

where matrix $U$ and vector $v$ are dual variables corresponding to constraints $R = H$ and $\beta = g$, respectively. Hyperparameters $\mu_1$ ad $\mu_2$ control the penalty of violation from the abovementioned constraints. $\mathcal{P}_\Omega(.)$ is the projection of its arguments on the space of non-empty elements. Also, $\langle .,. \rangle$ is the operator of the inner product in the Euclidean space. In order to minimize the augmented Lagrangian function in Equation (3), we need to consider the iteratively alternating algorithm between dual and primal optimizations. Equations (4) and (5) represent an alternating update procedure of primal and dual variables updating.

$$\text{Primal:} \quad (R^{k+1}, \beta^{k+1}, H^{k+1}, g^{k+1}) = \underset{R, \beta, H, g}{\operatorname{argmin}} \mathcal{L}(W^k, R, \beta, H, g, U^k, v^k), \tag{4}$$

$$\text{Dual:} \quad \begin{aligned} U^{k+1} &= U^k + \mu_1(R^{k+1} - H^{k+1}), \\ v^{k+1} &= v^k + \mu_2(\beta^{k+1} - g^{k+1}). \end{aligned} \tag{5}$$

Based on [16], we know that both L-1 and nuclear norm terms are convex [7,47]. In general, the first part of the loss function (2) and Equation (3) is non-convex since we assume that the weight matrix $W$ is iteratively updated by the sigmoid transformation. We have validated that implementing the block-coordinate updating by the ADMM solving procedure, i.e., updating few blocks of variables, [54] will lead us to a good local minima solution for this non-convex problem. Therefore, in this paper, we will use ADMM [5] with block updating to solve Equation (3). The key idea comes from (*i*) fixing weights at the beginning of each iteration so that we can relax the problem in each iteration to a convex problem; (*ii*) exploiting ADMM to find $\beta$ and $R$ for current iteration with the fixed weight; (*iii*) calculating the predicted organ-patient response matrix for the current iteration; and (iv) updating weight matrix of the next iteration based on the calculated prediction and sigmoid transformation. To sum up, in each iteration, we relax the bias term to the convex approximate by considering weights to be fixed when solving Equations (6)-(9), and perform the block updating. A solution can be obtained after converging to a locally optimal solution as shown in [20,50].

We further derive the subproblems for ADMM by breaking down Equation (4) into separate optimization problems corresponding to each unknown variable in Equations (6)-(9). Then, we implement the iteratively alternating minimization strategy through these smaller optimization problem solutions. The goal of using ADMM is to decouple the non-differentiable terms of Equation (3), i.e., $\|H\|_*$ and $\|g\|_1$, as separate problems and solve them by iterative alternating minimization to obtain $R$, $\beta$, $H$, and $g$. The alternating separated minor optimization problems based on each variable in Equation (4) will be formulated as follows.

$$\begin{aligned} R^{k+1} = \underset{R}{\operatorname{argmin}} \frac{1}{2} &\left\| \left((W^k)^{\frac{1}{2}} \odot \left(\mathcal{P}_\Omega(Y) - \mathcal{P}_\Omega(\mathcal{A}(X\beta^k))\right)\right) - \mathcal{P}_\Omega(R) \right\|_F^2 \\ &+ \langle U^k, R - H^k \rangle + \frac{\mu_1}{2} \|R - H^k\|_F^2, \end{aligned} \tag{6}$$

$$\begin{aligned} \beta^{k+1} = \underset{\beta}{\operatorname{argmin}} \frac{1}{2} &\left\| \left((W^k)^{\frac{1}{2}} \odot \left(\mathcal{P}_\Omega(Y) - \mathcal{P}_\Omega(\mathcal{A}(X\beta))\right)\right) - \mathcal{P}_\Omega(R^{k+1}) \right\|_F^2 \\ &+ \langle v^k, \beta - g^k \rangle + \frac{\mu_2}{2} \|\beta - g^k\|_F^2, \end{aligned} \tag{7}$$

$$H^{k+1} = \underset{H}{\operatorname{argmin}} \lambda_1 \|H\|_* + \langle U^k, R^{k+1} - H \rangle + \frac{\mu_1}{2} \|R^{k+1} - H\|_F^2, \tag{8}$$

$$g^{k+1} = \underset{g}{\operatorname{argmin}} \lambda_2 \|g\|_1 + \langle v^k, \beta^{k+1} - g \rangle + \frac{\mu_2}{2} \|\beta^{k+1} - g\|_2^2, \tag{9}$$

where Equations (6) and (7) are quadratic programming (QP) optimization problems. Therefore, the optimal solutions of $R$ and $\beta$ can be easily obtained by KKT First Order Necessary Conditions. For more details, see Appendix A.1 and A.2. However, because of non-differentiable terms, Equations (8) and (9) cannot be solved in the same way. Singular value soft-thresholding and wavelet soft-thresholding shrinkage will be used to solve Equations (8) and (9), respectively.

In the following paragraphs, we assume that $w \in [0,1]^{mn \times 1}$ is the reshaped vector from matrix $W \in [0,1]^{m \times n}$. We define $\mathbb{W}_{mn \times mn} = \{\operatorname{diag}(w), w = \operatorname{vec}(W) \in [0,1]^{mn \times 1}\}$ to simplify our calculations by transferring the element-wise multiplication to the usual matrix product. We also define $r \in \mathbb{R}^{mn \times 1}$, $y \in \mathbb{R}^{mn \times 1}$, $u \in \mathbb{R}^{mn \times 1}$ and $h \in \mathbb{R}^{mn \times 1}$ as corresponding vectors to matrices $R \in \mathbb{R}^{m \times n}$, $Y \in \mathbb{R}^{m \times n}$, $U \in \mathbb{R}^{m \times n}$ and $H \in \mathbb{R}^{m \times n}$, respectively.

By not considering $\|g\|_1$ in the objective function of Equation (9), we find $(\widehat{g}^{OLS})^{k+1} = \beta^{k+1} + \frac{v^k}{\mu_2}$. Then, the soft-thresholding shrinkage operator, denoted by $\mathcal{S}(.)$, will be used to solve Equation (9). Check Appendix A.3 for the elaborated definition of soft-thresholding shrinkage operator, i.e., $\mathcal{S}(.)$. Based on [6], singular value thresholding is proved to be the solution to the nuclear norm minimization problem. As with Equation (9), by not considering term $\|R\|_*$ in Equation (8), we find $(\widehat{H}^{OLS})^{k+1} = R^{k+1} + \frac{U^k}{\mu_1}$. Singular value thresholding operator, denoted by $\mathcal{D}(.)$, helps to solve Equation (7). See Appendix A.4 for the detailed definition of singular value thresholding operator, i.e., $\mathcal{D}(.)$. The solution of Equations (8) and (9) by exploiting $\mathcal{S}(.)$ and $\mathcal{D}(.)$ are derived as follows.

$$g^{k+1} = \mathcal{S}\left((\widehat{g}^{OLS})^{k+1}; \frac{\lambda_2}{\mu_2}\right) = \mathcal{S}\left(\beta^{k+1} + \frac{v^k}{\mu_2}; \frac{\lambda_2}{\mu_2}\right), \tag{10}$$

$$H^{k+1} = \mathcal{D}\left((\widehat{H}^{OLS})^{k+1}; \frac{\lambda_1}{\mu_1}\right) = \mathcal{D}\left(R^{k+1} + \frac{U^k}{\mu_1}; \frac{\lambda_1}{\mu_1}\right). \tag{11}$$

Then, after solving Equation (4) by ADMM, dual variables (i.e., $U^{k+1}$ and $v^{k+1}$) are attained based on updated primal variables (i.e., $R^{k+1}$, $\beta^{k+1}$, $H^{k+1}$, $g^{k+1}$) and alternating update of duals in Equation (5). Finally, the weight matrix will be updated based on the sigmoid transformation of the recommendation in the previous iteration.

$$\widehat{y}^k = -\left((\mathbb{W}^k)^{\frac{1}{2}}\left(y - (X\beta^k)\right) - r^k\right) + y, \tag{12}$$

$$w^{k+1} = \sigma(\widehat{y}^k) = \frac{1}{1 + e^{-\widehat{y}^k}}, \tag{13}$$

where $w^{k+1} \in [0,1]^{mn \times 1}$, $\widehat{y}^k \in \mathbb{R}^{mn \times 1}$, and $W^{k+1} = \mathcal{A}(w^{k+1})$, i.e., $W^{k+1} \in \mathbb{R}^{m \times n}$. The diagonal weight matrix also needs to be updated, $\mathbb{W}^{k+1}$, based on $W^{k+1}$.

It's important to note that the diagonal weight matrix can be initialized as an identity matrix $\mathbb{W}^0 = I_{mn \times mn}$, where the weight corresponds to all entries of the organ-patient response matrix are equal. However, to improve the convergence speed of the solving algorithm, we can also initialize the weight matrix from the output of the PRIME predicted response matrix [9]. The final solving procedure of optimization problem (2) is written in Algorithm 1. In this study, tolerance of convergence in Algorithm 1, i.e., $tol$, was set to be $10^{-6}$ to balance the speed and the accuracy. The tolerance is adjustable and can be changed by different needs in other studies.

**ALGORITHM 1: ADMM with Block Updating for AWTR**

**Inputs:** sparse organ-patient response matrix ($Y$), covariate matrix ($X$), tunning parameters $\lambda_1, \lambda_2, \mu_1$ and $\mu_2$
**Outputs:** estimated coefficient vector $\boldsymbol{\beta}$, low-rank matrix $R$, weight matrix $W$ and complete predicted matrix $\hat{Y}$
**Initialization:** $\mathbb{W}^0 \leftarrow diag(\boldsymbol{w}^0)$; $\boldsymbol{w}^0 \leftarrow \sigma(\hat{\boldsymbol{y}}^{PRIME})$; $\hat{\boldsymbol{y}}^{PRIME} \leftarrow \hat{\boldsymbol{r}}^{PRIME} + (X\hat{\boldsymbol{\beta}})^{PRIME}$;
$R^0, H^0, U^0 \leftarrow \boldsymbol{0}_{m \times n}$; $\boldsymbol{\beta}^0, \boldsymbol{g}^0, \boldsymbol{v}^0 \leftarrow \boldsymbol{0}_{p \times 1}$

**repeat**

$\hat{\boldsymbol{y}}^k \leftarrow -\left((\mathbb{W}^k)^{\frac{1}{2}}\left(\boldsymbol{y} - (X\boldsymbol{\beta}^k)\right) - \boldsymbol{r}^k\right) + \boldsymbol{y}$,

$\boldsymbol{r}^{k+1} \leftarrow \frac{1}{1+\mu_1}\left((\mathbb{W}^k)^{\frac{1}{2}}(\boldsymbol{y} - X\boldsymbol{\beta}^k) - \boldsymbol{u}^k + \mu_1 \boldsymbol{h}^k\right)$,

$\boldsymbol{\beta}^{k+1} \leftarrow \left(\left((\mathbb{W}^k)^{\frac{1}{2}}X\right)^T\left((\mathbb{W}^k)^{\frac{1}{2}}X\right) + \mu_2 I_{p \times p}\right)^{-1}\left\{\left((\mathbb{W}^k)^{\frac{1}{2}}X\right)^T\left((\mathbb{W}^k)^{\frac{1}{2}}\boldsymbol{y} - \boldsymbol{r}^{k+1}\right) - \boldsymbol{v}^k + \mu_2 \boldsymbol{g}^k\right\}$,

$H^{k+1} \leftarrow \mathcal{D}\left(R^{k+1} + \frac{U^k}{\mu_1}; \frac{\lambda_1}{\mu_1}\right)$,

$\boldsymbol{g}^{k+1} \leftarrow \mathcal{S}\left(\boldsymbol{\beta}^{k+1} + \frac{\boldsymbol{v}^k}{\mu_2}; \frac{\lambda_2}{\mu_2}\right)$,

$\boldsymbol{w}^{k+1} \leftarrow \sigma(\hat{\boldsymbol{y}}^k) = \frac{1}{1+e^{-\hat{\boldsymbol{y}}^k}}$,

$\mathbb{W}^{k+1} \leftarrow diag(\boldsymbol{w}^{k+1})$,

$U^{k+1} \leftarrow U^k + \mu_1(R^{k+1} - H^{k+1})$,

$\boldsymbol{v}^{k+1} = \boldsymbol{v}^k + \mu_2(\boldsymbol{\beta}^{k+1} - \boldsymbol{g}^{k+1})$,

**until:**

convergence: $\dfrac{\left\|\left((\mathbb{W}^{k+1})^{\frac{1}{2}}(\boldsymbol{y} - (X\boldsymbol{\beta}^{k+1}))\right) - \boldsymbol{r}^{k+1}) - \left((\mathbb{W}^k)^{\frac{1}{2}}(\boldsymbol{y} - (X\boldsymbol{\beta}^k)) - \boldsymbol{r}^k\right)\right\|_2^2}{\left\|\left((\mathbb{W}^k)^{\frac{1}{2}}(\boldsymbol{y} - (X\boldsymbol{\beta}^k)) - \boldsymbol{r}^k\right)\right\|_2^2} \leq tol$.

## 4 Simulation Case Study

In this simulation case study, we first illustrate the effectiveness of AWTR in the recommendation of top-N ranked patients, and then compare the performance of AWTR with other state-of-the-art top-N recommendation benchmarks. Simulation results will demonstrate that AWTR outperforms all the competing models in terms of top-N accuracy and top-N ranking accuracy when the initial organ-patient response matrix is highly sparse ($Y$).

We generate the covariate matrix ($X$) and the organ-patient response matrix ($Y$) with different levels of missing entries (sparsity) as inputs of the AWTR model. Missing entries will be positioned randomly in the organ-patient response matrix ($Y$) based on the sparsity level, i.e., the percentage of unobserved entries, where sparsity level ∈ {0.5, 0.7, 0.9, 0.95, 0.99}. The covariate matrix ($X$) will be generated based on all the normalized pairwise interactions between donors and patients' features. In kidney transplantation, each patient on the waiting list will have features (i.e., patient's age, BMI, blood type, gender, ethnicity, waiting time, albumin level, cause of the end-stage renal disease (ESRD), history of hypertension, history of diabetes, history of prior transplantation), and each organ will have features (i.e., donor's age, BMI, blood type, gender, ethnicity, acute kidney injury, serum creatinine level, cause of death, eGFR at donation, history of hypertension, history of diabetes). In this study, we consider 11 features for patients, 11 features for organs, and 121 features for the pairwise interactions of all organ-patient features and 1 feature for the distance of organ and patient geographical locations. Therefore, the total number of features or number of columns in the covariate matrix ($X$) is $p = 11 + 11 + 11^2 + 1 = 144$. Table 1 represents all the patient's and donor's features as well as their approximate distribution based on [41,53,60].

**Table 1: Feature descriptions [41,53,60]**

| Patient's Features | | Donor's Features | |
|---|---|---|---|
| **Features** | **Distribution** | **Features** | **Distribution** |
| Age (years) | Normal($\mu$= 46.56, $\sigma$= 12.66) | Age (years) | Normal($\mu$= 32.4, $\sigma$= 13.8) |
| Body Mass Index ($kg/m^2$) | Normal($\mu$= 25.91, $\sigma$= 5.46) | Body Mass Index ($kg/m^2$) | Normal($\mu$= 25.7, $\sigma$= 4.3) |
| Blood type | O45%, A40%, B11%, AB4% | Blood type | O45%, A40%, B11%, AB4% |
| Gender | Bernouli (M62% , F38%) | Gender | Bernouli(M:60% , F:40%) |
| Ethnicity | White:35.4%,African-American:32.1%, Hipanic:20.8%, Asian:9.2%, Others:2.5% | Ethnicity | White:65.5%, African American:15.1%, Hispanic:14.9%, Asian:2.6%, Others:1.9% |
| Waiting time (years) | Normal($\mu$= 3.2, $\sigma$= 2.47) | Acute kidney injury (AKI) | Bernouli( Yes:17.1% , No:82.9%) |
| Albumin level (g/DL) | Normal($\mu$= 3.83, $\sigma$= 0.72) | Serum Createnine level (mg/DL) | Normal($\mu$= 1.05, $\sigma$= 0.6) |
| Cause of ESRD | Glomerulonephritis:39%, Genetic kidney disease:14.5%, Diabetes:13.8%, Renovascular/hypertension:8%, Pyelonephritis:5.3%, Others:19.4% | Cause of death | Anoxia:10%, Cerebrovascular accident:31%, Central nervous system tumor:1%, Head trauma:47%, Others:11% |
| History of hypertension | Bernouli( Yes:16% , No:84%) | eGFR (mL/min) | Normal($\mu$= 83.1, $\sigma$= 31.2) |
| History of diabetes | Bernouli( Yes:37% , No:63%) | History of hypertension | Bernouli( Yes:14% , No:86%) |
| History of prior transplantation | Bernouli( Primary:88% , Repeat:12%) | History of diabetes | Bernouli( Yes:6% , No:94%) |

In the generated sparse organ-patient response matrix ($Y$), each row and column refers to an individual organ and patient, respectively. The number of organs is $m = 200$ and the number of patients is $n = 1000$. Based on the literature, there are multiple ways to estimate the proxy of post-transplant matching performance for kidneys such as Life Years from Transplant (LYFT) [52,53], Estimated Post Transplant Survival (EPTS) [65], and estimated patient's post-transplant Quality of Life (QoL) [32]. In this simulation study, we will use the KAS [60] to simulate entries of the organ-patient response matrix ($Y$). Based on [34,60], the KAS can be formulated as follows.

$$Y_{i,j} = KAS_{i,j} = 0.8 \times LYFT_{i,j} \times (1 - KDPI_i) + 0.8 \times DT_j \times KDPI_i$$
$$+ 0.2 \times DT_j + 0.04 \times CPRA_j, \quad i \in \{1,2,\ldots,m\}, \quad j \in \{1,2,\ldots,n\} \quad (14)$$

where the first two components are the life years from transplant (LYFT) and the patient's dialysis time (DT) scaled by the kidney donor profile index (KDPI) [66]. The third component only considers the dialysis time, and the fourth component considers the patient's antigen sensitivity by the calculated panel reactive antibodies (CPRA) [59]. Index $i$ and $j$ refer to the the $i$-th organ and the $j$-th patient, respectively. KAS integrates the effects of approximated post-transplant matching performances such as LYFT as well as patient's waiting time (Dialysis Time (DT)) and the approximated kidney quality (Kidney Donor Profile Index (KDPI) [66]) in the response data of this recommendation. In this simulation study, we leverage LYFT as one of the approximate post-transplant matching performance estimations. However, only considering LYFT as the simulated organ-patient response will not consider the effect of patient's waiting time and fairness in the organ matching decision-making. Also, KAS is preferred over LYFT since it combines the LYFT with other factors, i.e., it demonstrates response in a more complex way when real data is not available. That's why we selected KAS to generate the organ-patient response. Since entries of the organ-patient response matrix will be generated based on Equation (14), similar responses come from similar organs (rows) or similar patients (columns). Also, in practice, the generated organ-patient response matrix ($Y$) always has $rank < min\{m, n\} = m = 200$. Therefore, the low-rank structure holds in the organ-patient response matrix ($Y$) and the AWTR assumption is validated.

In this study, we focus on improving the top-N accuracy and the top-N ranking accuracy rather than overall ranking accuracy. To evaluate the top-N accuracy and the top-N ranking accuracy, we adopted two popular evaluation metrics as Hit Rate (HR) [14,30] and Normalized Discounted Cumulative Gain (NDCG) [12,49], respectively. HR is defined as follows.

$$HR = \frac{\# \, hits}{\# rows} = \frac{\frac{1}{N}\sum_{i=1}^{m} h_i}{\# rows}, \quad (15)$$

where $h_i = |T_i \cap \hat{T}_i|$ is the number of patients in $T_i$ that is also included in $\hat{T}_i$; $T_i \in \mathbb{R}^{1 \times N}$ and $\hat{T}_i \in \mathbb{R}^{1 \times N}$, $\forall i \in \{1,2,3,\ldots,m\}$, are the top-N list of ground truth and recommended patients for the $i$-th organ, respectively. If the HR is 1, then the proposed model can always include the true top-N patients in the top-N recommendation (i.e., the prediction from the model), which implies the best recommendation performance; if the HR is 0, then the model cannot recommend any of the true top-N patients in the top-N list, which implies the worst recommendation performance. The main limitation of HR is that it only considers the percentage of including real top-N patients in the recommendation, but it does not consider the rank of each patient in the top-N list. However, the exact ranking of patients in the top-N list is extremely important in the organ matching decision since the organ will be offered to patients based on their predicted ranking. The NDCG metric addresses this drawback [49]. The DCG and NDCG of each organ are defined as follows.

$$DCG_i = \sum_{z=1}^{N} \frac{2^{r_{i,z}} - 1}{log_2(z+1)},$$
$$NDCG_i = \frac{DCG_i}{IDCG_i}, \quad i \in \{1,2,\ldots,m\} \quad (16)$$

where $r_{i,z} \in \{0,1\}, \forall z \in \{1, 2, \ldots, N\}$ is the binary variable for the $z$-th element of the top-N patients corresponding to the $i$-th organ. If the position of $z$-th patient in the top-N list has been predicted accurate, i.e., $T_{i,z} = \hat{T}_{i,z}$, then $r_{i,z}$ takes the value 1. Otherwise, $r_{i,z}$ will be 0. Also, the $IDCG_i$ is the ideal $DCG_i$, which means the ranking of all top-N patients for the $i$-th organ is predicted correct, i.e., $r_{i,z} = 1$ for all $z \in \{1, 2, \ldots, N\}$. Finally, the NDCG of the whole recommendation matrix will be calculated by the average of $NDCG_i$ over all rows. When N = 1, HR and NDCG report the same value because there is only one position in the top-N list.

In order to investigate the sensitivity analysis of AWTR to correlation structures (i.e., figure out the impact of covariates' correlation on the performance of the AWTR), we performed simulation studies with different scenarios of correlation. We defined seven scenarios. In the first three scenarios, we will generate all the patients' and donors' covariates based on the multivariate normal distribution $N(\mathbf{0}, \mathbf{\Sigma})$ instead of the univariate distributions. In the first three scenarios, $\mathbf{\Sigma}$ is the covariance matrix which is set to be $\mathbf{\Sigma} = (\rho^{|i-j|})_{p \times p}$ with $\rho^{|i-j|}$ as the element in the $i$-th row and the $j$-th column and the correlation parameter $\rho \geq 0$ inspired by [26]. Then, we will use link functions [21,29] to transfer latent Gaussian variables and consider distributions of Table 1. For example, to transfer latent Gaussian variables to Bernoulli distributed variables, we will use the logit link function and then binarize the results based on the probability of success for the specific Bernoulli distribution. These three scenarios will consider low, moderate, and high correlation of covariates with $\rho$ taking values 0, 0.5, and 0.8, respectively. Table 2 and Table 3 represent the results of our simulation study for these three correlation scenarios. Based on the results, we can see that the performance of AWTR is robust in terms of top-N accuracy (HR) and top-N ranking accuracy (NDCG) for different levels of correlation among covariates. See Figure A.1 and Figure A.2 in Apendix A.6 for the correlation matrices of patients' and donors' covariates of low, moderate, and high correlations scenarios.

Table 2: HR performance of AWTR for low, moderate, and high correlation scenarios

| $\rho$ | N | HR | | | | |
|---|---|---|---|---|---|---|
| | | Sparsity = 50% | Sparsity = 70% | Sparsity = 90% | Sparsity = 95% | Sparsity = 99% |
| 0 | 1 | 0.935 | 0.811 | 0.506 | 0.345 | 0.290 |
| | 2 | 0.964 | 0.824 | 0.609 | 0.465 | 0.379 |
| | 5 | 0.974 | 0.910 | 0.626 | 0.517 | 0.472 |
| | 10 | 0.983 | 0.912 | 0.681 | 0.605 | 0.520 |
| 0.5 | 1 | 0.931 | 0.808 | 0.508 | 0.341 | 0.296 |
| | 2 | 0.948 | 0.845 | 0.601 | 0.471 | 0.366 |
| | 5 | 0.968 | 0.896 | 0.615 | 0.504 | 0.467 |
| | 10 | 0.971 | 0.904 | 0.686 | 0.601 | 0.506 |
| 0.8 | 1 | 0.932 | 0.804 | 0.502 | 0.336 | 0.281 |
| | 2 | 0.936 | 0.888 | 0.598 | 0.459 | 0.361 |
| | 5 | 0.952 | 0.876 | 0.613 | 0.509 | 0.478 |
| | 10 | 0.979 | 0.912 | 0.691 | 0.614 | 0.531 |

Table 3: NDCG performance of AWTR for low, moderate, and high correlation scenarios

| $\rho$ | N | NDCG | | | | |
|---|---|---|---|---|---|---|
| | | Sparsity = 50% | Sparsity = 70% | Sparsity = 90% | Sparsity = 95% | Sparsity = 99% |
| 0 | 1 | 0.935 | 0.811 | 0.506 | 0.345 | 0.290 |
| | 2 | 0.706 | 0.642 | 0.482 | 0.288 | 0.223 |
| | 5 | 0.548 | 0.384 | 0.264 | 0.219 | 0.207 |
| | 10 | 0.320 | 0.233 | 0.167 | 0.156 | 0.147 |
| 0.5 | 1 | 0.931 | 0.808 | 0.508 | 0.341 | 0.296 |
| | 2 | 0.702 | 0.633 | 0.485 | 0.284 | 0.219 |
| | 5 | 0.541 | 0.386 | 0.251 | 0.211 | 0.201 |
| | 10 | 0.324 | 0.238 | 0.162 | 0.168 | 0.151 |
| 0.8 | 1 | 0.932 | 0.804 | 0.502 | 0.336 | 0.281 |
| | 2 | 0.711 | 0.639 | 0.477 | 0.279 | 0.228 |
| | 5 | 0.535 | 0.378 | 0.248 | 0.196 | 0.182 |
| | 10 | 0.317 | 0.237 | 0.155 | 0.175 | 0.130 |

Since in this problem, we may face some covariates being highly correlated while they are not correlated with other covariates, we decided to define four new scenarios for block correlation structure. In these four scenarios, similarly, we will generate all the patients' and donors' covariates based on the multivariate Gaussian distribution $N(\mathbf{0}, \mathbf{\Sigma})$ and convert them to the distributions of Table 1 with proper link functions. The covariance matrix $\mathbf{\Sigma}$ will be scaled such that the largest elements in $\mathbf{\Sigma}$ take different values of $\varphi$ where $\varphi = 1, 1.8, 2.6,$ and $3.4$ refer to the four correlation scenarios with block correlation structures. Based on the results in Table 4 and Table 5, we can see that the performance of AWTR over different sparsity levels is also quite robust for these four block structure correlation. You can check Figure A.3 and Figure A.4 in Appendix A.6 for the correlation matrices of patients' and donors' covariates of these four block structure correlation scenarios.

**Table 4: HR performance of AWTR for different block structure correlation scenarios**

| $\varphi$ | N | HR | | | | |
|---|---|---|---|---|---|---|
| | | Sparsity = 50% | Sparsity = 70% | Sparsity = 90% | Sparsity = 95% | Sparsity = 99% |
| 1 | 1 | 0.931 | 0.816 | 0.502 | 0.348 | 0.281 |
| | 2 | 0.954 | 0.843 | 0.608 | 0.461 | 0.372 |
| | 5 | 0.972 | 0.915 | 0.635 | 0.523 | 0.490 |
| | 10 | 0.980 | 0.916 | 0.677 | 0.602 | 0.537 |
| 1.8 | 1 | 0.926 | 0.808 | 0.491 | 0.319 | 0.275 |
| | 2 | 0.982 | 0.844 | 0.588 | 0.468 | 0.341 |
| | 5 | 0.986 | 0.901 | 0.663 | 0.501 | 0.484 |
| | 10 | 0.991 | 0.926 | 0.685 | 0.564 | 0.514 |
| 2.6 | 1 | 0.921 | 0.821 | 0.508 | 0.331 | 0.265 |
| | 2 | 0.991 | 0.851 | 0.596 | 0.454 | 0.338 |
| | 5 | 0.994 | 0.918 | 0.688 | 0.537 | 0.457 |
| | 10 | 0.993 | 0.927 | 0.686 | 0.545 | 0.528 |
| 3.4 | 1 | 0.942 | 0.805 | 0.501 | 0.326 | 0.278 |
| | 2 | 0.952 | 0.878 | 0.628 | 0.477 | 0.349 |
| | 5 | 0.978 | 0.934 | 0.696 | 0.582 | 0.477 |
| | 10 | 0.984 | 0.937 | 0.703 | 0.613 | 0.506 |

**Table 5: NDCG performance of AWTR for different block structure correlation scenarios**

| $\varphi$ | N | NDCG | | | | |
|---|---|---|---|---|---|---|
| | | Sparsity = 50% | Sparsity = 70% | Sparsity = 90% | Sparsity = 95% | Sparsity = 99% |
| 1 | 1 | 0.931 | 0.816 | 0.502 | 0.348 | 0.281 |
| | 2 | 0.701 | 0.651 | 0.491 | 0.278 | 0.227 |
| | 5 | 0.531 | 0.376 | 0.269 | 0.212 | 0.202 |
| | 10 | 0.318 | 0.235 | 0.163 | 0.151 | 0.138 |
| 1.8 | 1 | 0.926 | 0.808 | 0.491 | 0.319 | 0.275 |
| | 2 | 0.682 | 0.646 | 0.483 | 0.269 | 0.201 |
| | 5 | 0.535 | 0.352 | 0.264 | 0.201 | 0.181 |
| | 10 | 0.304 | 0.290 | 0.165 | 0.164 | 0.114 |
| 2.6 | 1 | 0.921 | 0.821 | 0.508 | 0.331 | 0.265 |
| | 2 | 0.691 | 0.654 | 0.476 | 0.250 | 0.228 |
| | 5 | 0.523 | 0.341 | 0.287 | 0.236 | 0.213 |
| | 10 | 0.273 | 0.227 | 0.186 | 0.143 | 0.108 |
| 3.4 | 1 | 0.942 | 0.805 | 0.501 | 0.326 | 0.278 |
| | 2 | 0.712 | 0.672 | 0.491 | 0.279 | 0.244 |
| | 5 | 0.532 | 0.371 | 0.298 | 0.187 | 0.179 |
| | 10 | 0.281 | 0.228 | 0.153 | 0.144 | 0.102 |

To analyze the performance of AWTR in the top-N recommendation and top-N rank learning, we implemented AWTR on a simple simulation study where $Y_{10 \times 10}$ with 70% sparsity level, $X_{100 \times 20}$, and we consider the recommendation list size as N = 5. Figure 3 shows how the adaptation of the weight matrix over iterations will improve the recommendation of top-ranked patients. We will illustrate this improvement in three stages as initialization, middle iteration(s), and the final iteration. Each row refers to an individual organ and each column refers to a patient. The numbers in the matrix represent the ranking of patients' assignment to each organ based on their predicted matching performance. Figure 3 shows that AWTR will improve the top-N accuracy and top-N ranking accuracy of recommendations. For example, in the second row, the top-N accuracy (i.e., HR) will increase from 20% in the initialization to 100% in the final iteration. Similarly, the top-N ranking accuracy (i.e., NDCG) will increase from 21.3% in the initialization to 47% in the final iteration. It can be observed that the predicted ranking based on the AWTR in the final iteration has a good agreement with the ground truth ranking of top-5 patients corresponding to each organ although 70% of these historical data are missing in AWTR's input.

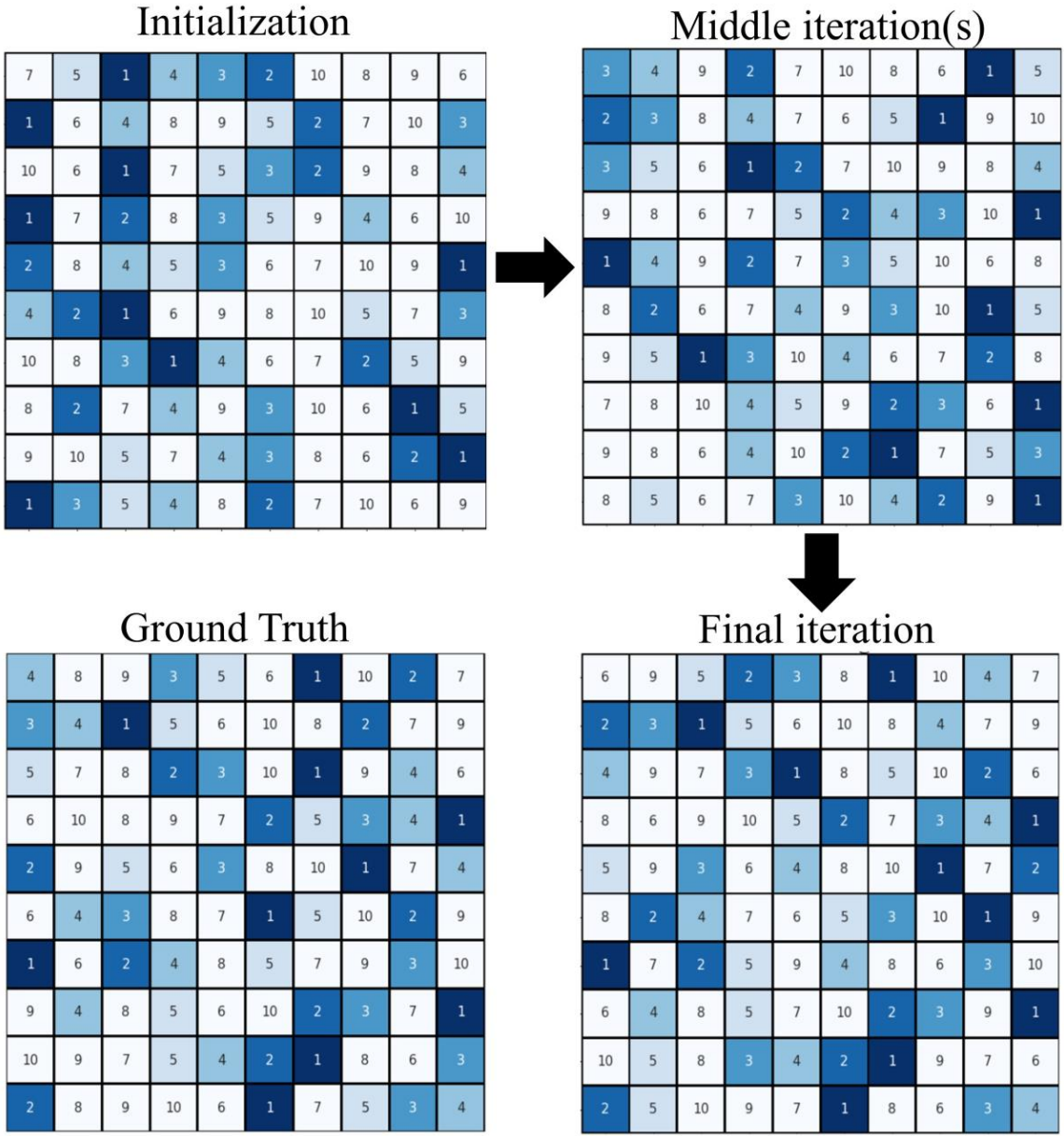

**Figure 3: An example of predictions based on different iterations of weight matrix improvement compared to the ground truth for top-5 patients in gradient colors. (columns: patients; rows: organs).**

We compare the performance of AWTR with seven state-of-the-art top-N recommendation methods. In this study, each method will be tested by using 50 replications, and the evaluation metric values are reported as the mean of all replications. We test the performance of all models for different sparsity levels, where sparsity level∈ {0.5,0.7,0.9,0.95,0.99}. Also, we define four possible scenarios for $N \in \{1, 2, 5, 10\}$ rather than only considering the accuracy of predicting the best-matched patient ($N = 1$) since the offered organ may be rejected by the best-matched patient. Based on [8], using pairwise loss instead of least square loss in the objective function of the recommender model will improve the robustness and top-N accuracy of the model. Because of this, we consider both PRIME [9] and PRIME with pairwise loss as two benchmarks. Besides, low-rank matrix completion (LorMC) [6,7,25] is a well-known recommendation model that performs the recommendation based on the low-rank property (i.e., implicit similarities among rows and columns). In this evaluation study, we also consider LorMC with pairwise loss as the third benchmark method [8], and we implement this method by using nuclear norm minimization in the objective function. In addition to these three benchmarks, sparse linear method (SLIM) [30] and low-rank sparse linear method (LorSLIM) [10] are two other top-N recommendation methods that concentrate on the top-N rankings and outperform the existing CF models in terms of top-N recommendation accuracy. Since deep learning recommendation models have recently attracted interests in the recommendation literature and have shown good performance in the top-N recommendation, we also consider neural collaborative filtering (NCF) [22]; and deep factorization machines (DeepFM) [19] as the last two benchmark methods in our study. Therefore, our comparison study is performed with seven benchmarks: (1) PRIME with least square loss; (2) PRIME with pairwise loss; (3) LorMC with pairwise loss; (4) SLIM; (5) LorSLIM; (6) NCF; and (7) DeepFM.

Figure 4 compares the HR of AWTR and all the seven benchmarks when $m = 200$ (see Table A.1 in Appendix A.5 for numerical results). Based on the results, it can be observed that when $m = 200$ and $n = 1000$, our proposed method (AWTR) outperforms benchmarks when the input data is highly sparse (sparsity $\geq 99\%$) which is the common case in the organ matching problem. The performance of NCF and DeepFM usually drop when sparsity > 95% which might happen due to the lack of sufficient data to fully support their rich parametrization. Since most of the deep learning methods are data-hungry, we also tried the simulation study with $m = 20$. Figure 5 represents the HR of AWTR and all the seven benchmarks when $m = 20$ (see Table A.1 in Appendix A.5 for numerical results). Based on the results, we can see that when $m = 20$ and $n = 1000$, AWTR outperforms all other seven benchmarks in more scenarios and especially when sparsity > 90%. Although the higher HR means the higher probability of containing top-N patients in the prediction, the higher HR will not ensure that the ranking of patients in each top-N list is always predicted accurately. Therefore, we also consider the NDCG metric in our comparison study. Figure 6 and Figure 7 compare the NDCG of AWTR and other benchmarks when $m = 200$ and when $m = 20$, respectively (see Table A.2 in Appendix A.5 for numerical results). Similar to the HR results, it can be observed that the performance of NCF and DeepFM is better than other methods when sparsity < 90% and their performance suddenly decreases when sparsity is high (sparsity > 90%) and sample size decreases ($m = 20$).

Based on the figures and numerical results, we can see that SLIM works great in both HR and NDCG when the sparsity level is low. However, as sparsity level increases, SLIM fails to learn ranking and encounters a sudden significant drop in terms of both top-N accuracy (i.e., HR) and top-N ranking accuracy (i.e., NDCG). This is because SLIM can only capture relations between patients that are co-offered by at least one organ. To overcome this limitation of SLIM, LorSLIM [10] was proposed by imposing a low-rank constraint in the SLIM loss function. This low-rank constraint makes SLIM more robust and reduces its sudden changes in the ranking accuracy. Numerical results show that the proposed AWTR is even more robust than LorSLIM when the sparsity level is high. Also, based on the figures and tables, it can be observed that NCF and DeepFM outperform SLIM when sparsity < 70%, and they can also outperform LorSLIM and AWTR when sparsity < 90%. However, as we explained in the previous paragraph, the performance of NCF and DeepFM suddenly drop, and AWTR outperforms both when the response data is highly sparse and the sample size is small. Therefore, the results illustrate effectiveness of the iterative weight matrix adaptation making AWTR performs better in terms of HR and NDCG. The iterative update of weights will improve the top-N recommendation and prevent AWTR from a sudden change in HR and NDCG when small proportion of historical data is available. In addition, NCF and DeepFM both require extensive hyperparameter tuning, and their performance is very sensitive to the tuning procedure, while AWTR has significantly less number of hyperparameters (only 2 in loss) and will automatically tune them in the efficient way. Also, based on the numerical results, HR almost increases with N, which implies that the probability of the true top-ranked patients being included in the predicted top-N recommendation will increase when N increases; and NDCG almost decreases with N, which implies that when N increases, the chance of predicting ranking accurately will decrease since there are more positions in the top-N list that need to be ranked correctly.

Finally, the L-1 regularized regression term in AWTR enables the variable selection by investigating the non-zero items in coefficients vector $\boldsymbol{\beta}$. Specifically, AWTR selected the (patient age)×(donor age); (patient BMI)×(donor BMI); (patient age)×(donor BMI); (patient BMI)×(donor age); (patient cause of ESRD)×(donor age); (patient history of diabetes)×(donor age); and (patient cause of ESRD)×(donor eGFR at donation) as seven significant features based on the interaction of patients' and donors' features for our simulated kidney matching decision-making problem.

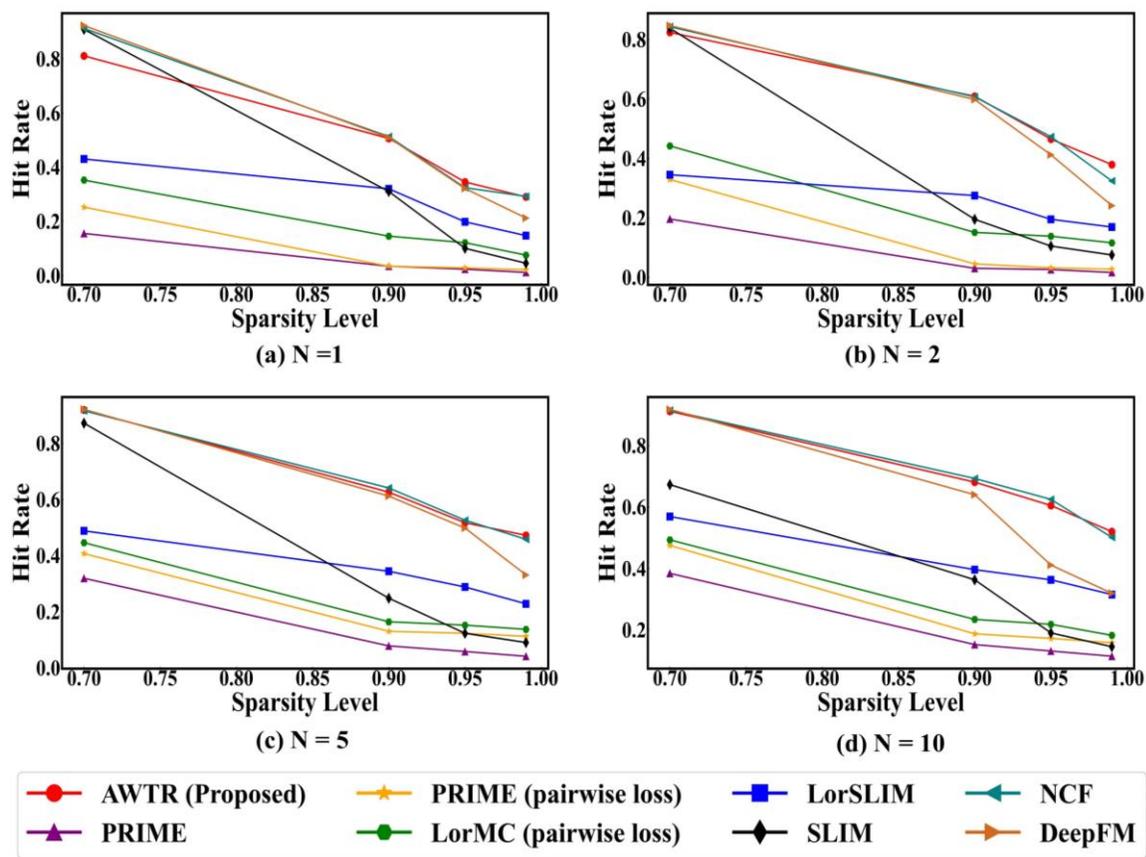

**Figure 4:** Comparison of HR performance for AWTR vs other benchmarks over sparsity levels greater than 70% and varying N ∈ {1, 2, 5, 10} with $m = 200, n = 1000$.

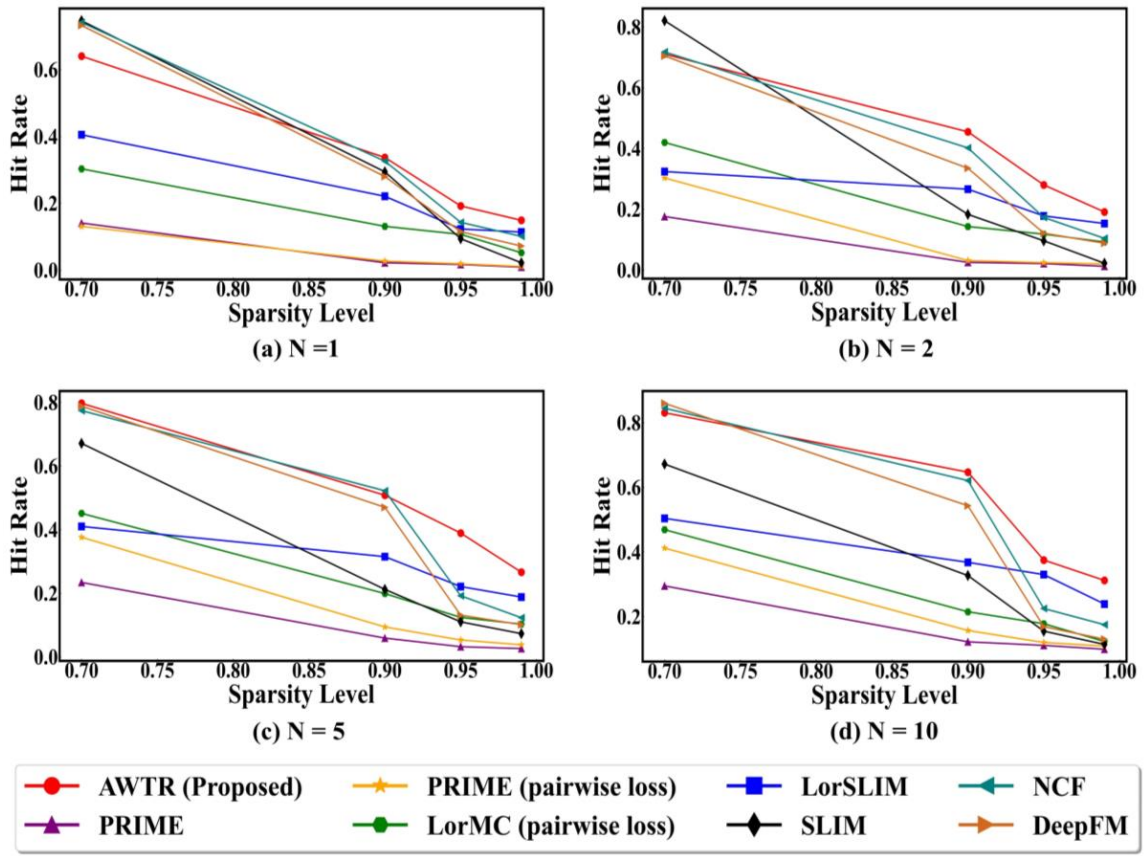

**Figure 5: Comparison of HR performance for AWTR vs other benchmarks over sparsity levels greater than 70% and varying N ∈ {1, 2, 5, 10} with $m = 20, n = 1000$.**

Figure 6: Comparison of NDCG performance for AWTR vs other benchmarks over sparsity levels greater than 70% and varying N ∈ {1, 2, 5, 10} with $m = 200, n = 1000$.

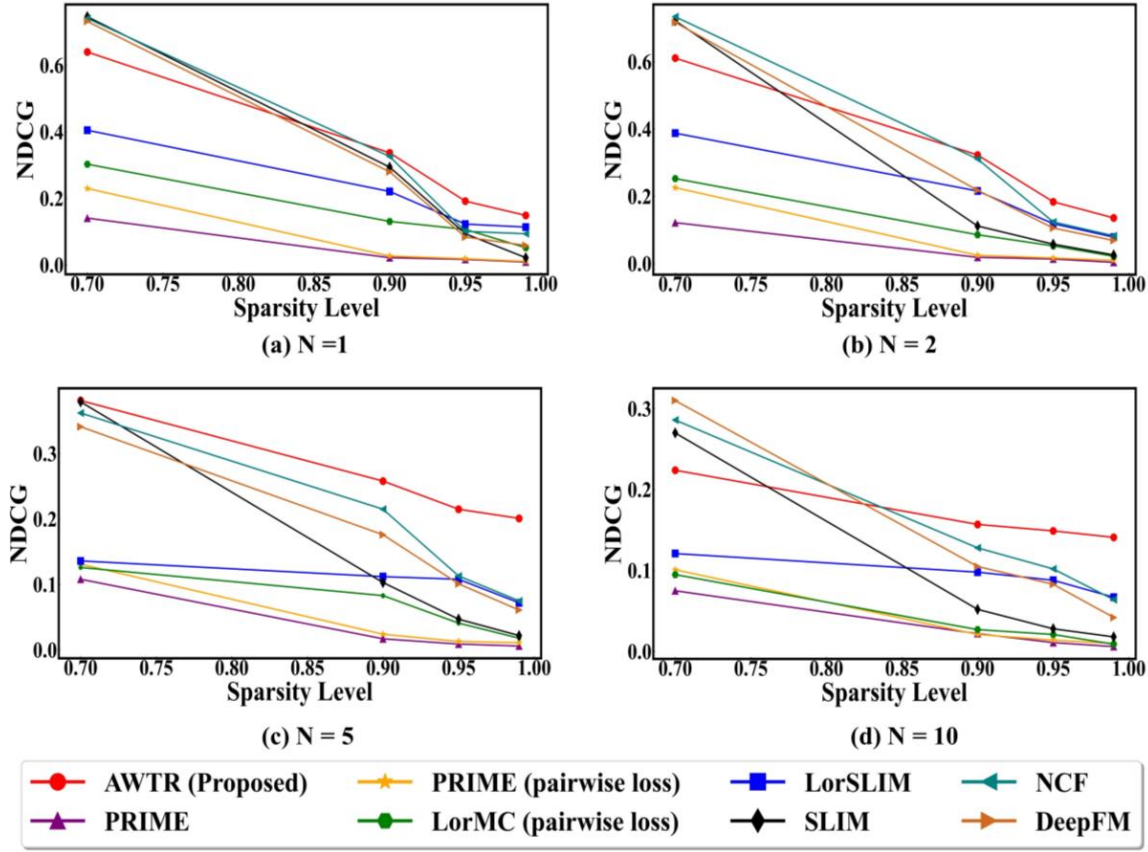

**Figure 7: Comparison of NDCG performance for AWTR vs other benchmarks over sparsity levels greater than 70% and varying N ∈ {1, 2, 5, 10} with $m = 20, n = 1000$.**

# 5 Conclusions

The organ matching is one of the most substantial problems in organ transplantation network management. The objective of this problem is the allocation of limited viable organs to the most "suitable" patients. In the current practice, organ matching decisions were only made by algorithmic scoring models based on the first principles in a deterministic manner. However, these scoring models may not be in line with the actual post-transplant matching performance. In this paper, we formulated organ matching as a top-N recommendation problem and proposed AWTR method. AWTR will improve performance of the current scoring system by using limited post-transplant matching performance in historical data set as well as the collected covariates from both organ donors and patients. AWTR concentrates on the predicted top-N patients and their ranking accuracy instead of accuracy of the predicted matching scores for all patients on the waiting list. In AWTR, the top-N matched patients will be iteratively recommended by adaptively adjusting the weight matrix in the loss function by the sigmoid transformation of the prediction in the latest iteration. The simulation case study showed that AWTR benefited from adaptive weights and outperformed seven top-N recommendation benchmark methods in terms of HR and NDCG when the sparsity level is high.

There are several future research directions. First, since our proposed recommendation model is able to learn the sparse coefficient vector of covariates (i.e., $\boldsymbol{\beta}$), we can select the most significant features which have been generated based on all pairwise interactions between covariates from organ donors and patients. One future work of this research is to include new features for organ-patient matching, whose significance can be identified from AWTR. It has a potential to improve the current matching scoring models and matching performance. In the meantime, some constraints such as donor-patient distance, and available resources (e.g., transplant surgeons) need to be considered in the organ matching decision. Therefore, we need to think about new optimization problems that can integrate these constraints into the recommendation model. We want to apply the proposed method on real data of the actual post-transplantation matching performance for validation.

## A APPENDICES

## A.1 Solving parameter *R* decoupled problem by KKT conditions

As in Equation (6), the decoupled optimization problem for the parameter $\boldsymbol{R}$ is as follows.

$$\boldsymbol{R}^{k+1} = \underset{\boldsymbol{R}}{argmin}\frac{1}{2}\left\|\left((\boldsymbol{W}^k)^{\frac{1}{2}}\odot\left(\mathcal{P}_\Omega(\boldsymbol{Y})-\mathcal{P}_\Omega\left(\mathcal{A}(\boldsymbol{X}\boldsymbol{\beta}^k)\right)\right)-\mathcal{P}_\Omega(\boldsymbol{R})\right)\right\|_F^2$$
$$+\langle \boldsymbol{U}^k, \boldsymbol{R}-\boldsymbol{H}^k\rangle + \frac{\mu_1}{2}\|\boldsymbol{R}-\boldsymbol{H}^k\|_F^2.$$

This unconstrained optimization problem can be solved by KKT conditions as shown.

$$f(\boldsymbol{r}) = \frac{1}{2}\left\|\left((\mathbb{W}^k)^{\frac{1}{2}}\cdot(\boldsymbol{y}-\boldsymbol{X}\boldsymbol{\beta}^k)-\boldsymbol{r}\right)\right\|_F^2 + \langle \boldsymbol{u}^k, \boldsymbol{r}-\boldsymbol{h}^k\rangle + \frac{\mu_1}{2}\|\boldsymbol{r}-\boldsymbol{h}^k\|_F^2, \quad (A1.a)$$

$$\frac{\partial f(\boldsymbol{r})}{\partial \boldsymbol{r}} = -((\mathbb{W}^k)^{\frac{1}{2}}\cdot(\boldsymbol{y}-\boldsymbol{X}\boldsymbol{\beta}^k)-\boldsymbol{r}) + \boldsymbol{u}^k + \mu_1(\boldsymbol{r}-\boldsymbol{h}^k) = 0, \quad (A1.b)$$

$$\boldsymbol{r}(1+\mu_1) = (\mathbb{W}^k)^{\frac{1}{2}}\cdot(\boldsymbol{y}-\boldsymbol{X}\boldsymbol{\beta}^k) - \boldsymbol{u}^k + \mu_1\boldsymbol{h}^k,$$

$$\boldsymbol{r}^{k+1} = \frac{1}{1+\mu_1}\left((\mathbb{W}^k)^{\frac{1}{2}}\cdot(\boldsymbol{y}-\boldsymbol{X}\boldsymbol{\beta}^k) - \boldsymbol{u}^k + \mu_1\boldsymbol{h}^k\right).$$

## A.2 Solving parameter *β* decoupled problem by KKT conditions

Based on Equation (7), the decoupled optimization problem for the parameter $\boldsymbol{\beta}$ is as follows.

$$\boldsymbol{\beta}^{k+1} = \underset{\boldsymbol{\beta}}{argmin}\frac{1}{2}\left\|(\boldsymbol{W}^k)^{\frac{1}{2}}\odot\left(\mathcal{P}_\Omega(\boldsymbol{Y})-\mathcal{P}_\Omega(\mathcal{A}(\boldsymbol{X}\boldsymbol{\beta}))\right)-\mathcal{P}_\Omega(\boldsymbol{R}^{k+1})\right\|_F^2$$
$$+\langle \boldsymbol{v}^k, \boldsymbol{\beta}-\boldsymbol{g}^k\rangle + \frac{\mu_2}{2}\|\boldsymbol{\beta}-\boldsymbol{g}^k\|_F^2.$$

This problem can be solved by KKT conditions as demonstrated.

$$f(\boldsymbol{\beta}) = \frac{1}{2}\left\|(\mathbb{W}^k)^{\frac{1}{2}}\cdot(\boldsymbol{y}-\boldsymbol{X}\boldsymbol{\beta})-\boldsymbol{r}^{k+1}\right\|_F^2 + \langle \boldsymbol{v}^k, \boldsymbol{\beta}-\boldsymbol{g}^k\rangle + \frac{\mu_2}{2}\|\boldsymbol{\beta}-\boldsymbol{g}^k\|_F^2, \quad (A2.a)$$

$$\frac{\partial f(\boldsymbol{\beta})}{\partial \boldsymbol{\beta}} = -((\mathbb{W}^k)^{\frac{1}{2}} \cdot \boldsymbol{X})^T \cdot ((\mathbb{W}^k)^{\frac{1}{2}} \cdot (\boldsymbol{y} - \boldsymbol{X}\boldsymbol{\beta}) - \boldsymbol{r}^{k+1}) + \boldsymbol{v}^k + \mu_2(\boldsymbol{\beta} - \boldsymbol{g}^k) = 0,$$

$$\boldsymbol{\beta}^{k+1} = \left(\left((\mathbb{W}^k)^{\frac{1}{2}} \cdot \boldsymbol{X}\right)^T \cdot \left((\mathbb{W}^k)^{\frac{1}{2}} \cdot \boldsymbol{X}\right) + \mu_2 I_{p \times p}\right)^{-1} \cdot \left[\left((\mathbb{W}^k)^{\frac{1}{2}} \cdot \boldsymbol{X}\right)^T \cdot \left((\mathbb{W}^k)^{\frac{1}{2}} \cdot \boldsymbol{y} - \boldsymbol{r}^{k+1}\right) - \boldsymbol{v}^k + \mu_2 \boldsymbol{g}^k\right]. \quad \text{(A2.b)}$$

## A.3 Lasso regression estimator based on soft-thresholding shrinkage

In a simple lasso regression problem, the lasso estimator can be found by soft-thresholding shrinkage operator as follows [16].

$$\hat{\theta}_j = \underset{\theta_j}{argmin} \frac{1}{2}\left(\hat{\theta}_j^{OLS} - \theta_j\right)^2 + \lambda|\theta_j|, \quad \text{(A3.a)}$$

$$\hat{\theta}_j = s(\hat{\theta}_j^{OLS}; \lambda), \quad \text{(A3.b)}$$

$$s(\hat{\theta}_j^{OLS}; \lambda) = sign(\hat{\theta}_j^{OLS})(|\hat{\theta}_j^{OLS}| - \lambda)_+ = \begin{cases} \hat{\theta}_j^{OLS} - \lambda & \text{if } \hat{\theta}_j^{OLS} > \lambda \\ 0 & \text{if } |\hat{\theta}_j^{OLS}| \leq \lambda \\ \hat{\theta}_j^{OLS} + \lambda & \text{if } \hat{\theta}_j^{OLS} < -\lambda \end{cases},$$

where $(|\hat{\theta}_j^{OLS}| - \lambda)_+ = max(0, |\hat{\theta}_j^{OLS}| - \lambda)$. Therefore, the soft-thresholding shrinkage operator of vector $\boldsymbol{\theta}$ will be defined as follows.

$$\mathcal{S}(\hat{\boldsymbol{\theta}}^{OLS}; \lambda) = [s\left(\hat{\theta}_1^{OLS}; \lambda\right), s\left(\hat{\theta}_2^{OLS}; \lambda\right), s\left(\hat{\theta}_3^{OLS}; \lambda\right), \ldots]. \quad \text{(A3.c)}$$

## A.4 Definition of singular value soft-thresholding operator

We consider the singular value decomposition (SVD) of matrix $\boldsymbol{A}$ such that $\boldsymbol{A}_{m \times n} = \boldsymbol{P}_{m \times t} \boldsymbol{\Sigma}_{t \times t} \boldsymbol{Q}^T_{t \times n}$ where $\boldsymbol{\Sigma} = \text{diag}(\{\sigma_i\}_{1 \leq i \leq t})$. Based on [6], singular value thresholding operator will be defined as follows.

$$\mathcal{D}(\boldsymbol{A}; \lambda) = \boldsymbol{P}\boldsymbol{\Sigma}'\boldsymbol{Q}^T, \quad \text{(A4.a)}$$

where $\boldsymbol{\Sigma}' = \text{diag}(\{\sigma_i - \lambda\}_{+\ 1 \leq i \leq t})$ and $\{\sigma_i - \lambda\}_+ = max(0, \sigma_i - \lambda)$. Singular value thresholding is proved to be the solution to the nuclear norm minimization problem [6].

## A.5 Evaluation Study Tables

Table A.1 and Table A.2 represent performance of the proposed model, AWTR, in comparison to other benchmarks in terms of HR and NDCG, respectively.

Table A.1: HR performance for AWTR vs other benchmarks over sparsity levels greater than 50% and varying N ∈ {1, 2, 5, 10}

| $m$ | N | Methodology | Hit Rate | | | | |
|---|---|---|---|---|---|---|---|
| | | | Sparsity = 50% | Sparsity = 70% | Sparsity = 90% | Sparsity = 95% | Sparsity = 99% |
| 200 | 1 | AWTR (Proposed) | 0.935 | 0.811 | 0.506 | **0.345** | 0.290 |
| | | LorSLIM | 0.541 | 0.430 | 0.320 | 0.198 | 0.147 |
| | | SLIM | 0.993 | 0.909 | 0.310 | 0.102 | 0.045 |
| | | NCF | 0.994 | 0.912 | **0.513** | 0.324 | **0.292** |
| | | DeepFM | **0.996** | **0.923** | 0.509 | 0.321 | 0.212 |
| | | LorMC (pairwise loss) | 0.602 | 0.352 | 0.145 | 0.121 | 0.075 |
| | | PRIME ( pairwise loss) | 0.517 | 0.252 | 0.034 | 0.028 | 0.021 |
| | | PRIME | 0.322 | 0.155 | 0.034 | 0.023 | 0.011 |
| | 2 | AWTR (Proposed) | 0.964 | 0.824 | **0.609** | 0.465 | **0.379** |
| | | LorSLIM | 0.585 | 0.345 | 0.275 | 0.195 | 0.169 |
| | | SLIM | 0.995 | 0.838 | 0.195 | 0.105 | 0.035 |
| | | NCF | 0.994 | 0.843 | 0.607 | **0.473** | 0.324 |
| | | DeepFM | **0.997** | **0.847** | 0.598 | 0.412 | 0.241 |
| | | LorMC (pairwise loss) | 0.652 | 0.442 | 0.151 | 0.138 | 0.116 |
| | | PRIME ( pairwise loss) | 0.545 | 0.329 | 0.045 | 0.032 | 0.027 |
| | | PRIME | 0.403 | 0.196 | 0.03 | 0.026 | 0.016 |
| | 5 | AWTR (Proposed) | 0.974 | 0.910 | 0.626 | 0.517 | **0.472** |
| | | LorSLIM | 0.619 | 0.488 | 0.344 | 0.288 | 0.228 |
| | | SLIM | 0.992 | 0.682 | 0.248 | 0.123 | 0.090 |
| | | NCF | **0.996** | 0.916 | **0.641** | **0.526** | 0.458 |
| | | DeepFM | 0.995 | **0.921** | 0.612 | 0.498 | 0.331 |
| | | LorMC (pairwise loss) | 0.679 | 0.446 | 0.164 | 0.152 | 0.137 |
| | | PRIME ( pairwise loss) | 0.621 | 0.407 | 0.130 | 0.123 | 0.112 |
| | | PRIME | 0.538 | 0.319 | 0.078 | 0.058 | 0.041 |
| | 10 | AWTR (Proposed) | 0.983 | 0.912 | 0.681 | 0.605 | **0.520** |
| | | LorSLIM | 0.728 | 0.569 | 0.396 | 0.363 | 0.315 |
| | | SLIM | 0.977 | 0.673 | 0.363 | 0.190 | 0.145 |
| | | NCF | 0.989 | 0.915 | **0.693** | **0.625** | 0.501 |
| | | DeepFM | **0.991** | **0.918** | 0.640 | 0.411 | 0.319 |
| | | LorMC (pairwise loss) | 0.706 | 0.493 | 0.234 | 0.218 | 0.182 |
| | | PRIME ( pairwise loss) | 0.671 | 0.475 | 0.187 | 0.172 | 0.158 |
| | | PRIME | 0.572 | 0.384 | 0.152 | 0.131 | 0.114 |
| 20 | 1 | AWTR (Proposed) | 0.811 | 0.640 | **0.337** | **0.192** | **0.149** |
| | | LorSLIM | 0.513 | 0.405 | 0.221 | 0.123 | 0.114 |
| | | SLIM | 0.871 | **0.746** | 0.294 | 0.094 | 0.022 |
| | | NCF | 0.878 | 0.741 | 0.326 | 0.143 | 0.101 |
| | | DeepFM | **0.894** | 0.732 | 0.280 | 0.115 | 0.072 |
| | | LorMC (pairwise loss) | 0.560 | 0.303 | 0.131 | 0.107 | 0.052 |
| | | PRIME ( pairwise loss) | 0.494 | 0.230 | 0.027 | 0.019 | 0.011 |
| | | PRIME | 0.306 | 0.141 | 0.022 | 0.017 | 0.009 |
| | 2 | AWTR (Proposed) | 0.834 | 0.711 | **0.455** | **0.280** | **0.191** |
| | | LorSLIM | 0.532 | 0.324 | 0.266 | 0.178 | 0.150 |
| | | SLIM | 0.920 | **0.821** | 0.183 | 0.096 | 0.022 |
| | | NCF | **0.923** | 0.718 | 0.402 | 0.173 | 0.104 |
| | | DeepFM | 0.917 | 0.705 | 0.335 | 0.121 | 0.088 |
| | | LorMC (pairwise loss) | 0.634 | 0.420 | 0.143 | 0.118 | 0.092 |
| | | PRIME ( pairwise loss) | 0.521 | 0.303 | 0.031 | 0.024 | 0.020 |
| | | PRIME | 0.382 | 0.176 | 0.025 | 0.021 | 0.012 |
| | 5 | AWTR (Proposed) | 0.888 | **0.797** | 0.508 | **0.389** | **0.266** |
| | | LorSLIM | 0.555 | 0.410 | 0.315 | 0.221 | 0.188 |
| | | SLIM | 0.944 | 0.671 | 0.212 | 0.110 | 0.073 |
| | | NCF | 0.951 | 0.774 | **0.522** | 0.192 | 0.123 |
| | | DeepFM | **0.959** | 0.788 | 0.470 | 0.131 | 0.101 |
| | | LorMC (pairwise loss) | 0.645 | 0.451 | 0.199 | 0.124 | 0.103 |
| | | PRIME ( pairwise loss) | 0.559 | 0.376 | 0.094 | 0.053 | 0.038 |
| | | PRIME | 0.427 | 0.234 | 0.059 | 0.032 | 0.026 |
| | 10 | AWTR (Proposed) | 0.949 | 0.831 | **0.647** | **0.375** | **0.312** |
| | | LorSLIM | 0.622 | 0.504 | 0.368 | 0.330 | 0.239 |
| | | SLIM | 0.957 | 0.672 | 0.327 | 0.155 | 0.114 |
| | | NCF | 0.966 | 0.845 | 0.621 | 0.225 | 0.175 |
| | | DeepFM | **0.972** | **0.860** | 0.543 | 0.169 | 0.130 |
| | | LorMC (pairwise loss) | 0.688 | 0.469 | 0.215 | 0.178 | 0.123 |
| | | PRIME (pairwise loss) | 0.626 | 0.412 | 0.157 | 0.120 | 0.109 |
| | | PRIME | 0.488 | 0.295 | 0.122 | 0.111 | 0.099 |

Table A.2: NDCG performance for AWTR vs other benchmarks over sparsity levels greater than 50% and varying $N \in \{1, 2, 5, 10\}$

| $m$ | N | Methodology | Normalized Discounted Cumulative Gain (NDCG) | | | | |
|---|---|---|---|---|---|---|---|
| | | | Sparsity = 50% | Sparsity = 70% | Sparsity = 90% | Sparsity = 95% | Sparsity = 99% |
| 200 | 1 | AWTR (Proposed) | 0.935 | 0.811 | 0.506 | **0.345** | 0.290 |
| | | LorSLIM | 0.541 | 0.430 | 0.320 | 0.198 | 0.147 |
| | | SLIM | 0.993 | 0.909 | 0.310 | 0.102 | 0.045 |
| | | NCF | 0.994 | 0.912 | **0.513** | 0.324 | **0.292** |
| | | DeepFM | **0.996** | **0.923** | 0.509 | 0.321 | 0.212 |
| | | LorMC (pairwise loss) | 0.602 | 0.352 | 0.145 | 0.121 | 0.075 |
| | | PRIME (pairwise loss) | 0.517 | 0.252 | 0.034 | 0.028 | 0.021 |
| | | PRIME | 0.322 | 0.155 | 0.034 | 0.023 | 0.011 |
| | 2 | AWTR (Proposed) | 0.706 | 0.642 | 0.482 | **0.288** | **0.223** |
| | | LorSLIM | 0.446 | 0.398 | 0.228 | 0.163 | 0.088 |
| | | SLIM | 0.992 | 0.745 | 0.123 | 0.062 | 0.028 |
| | | NCF | 0.994 | **0.769** | **0.502** | 0.273 | 0.194 |
| | | DeepFM | **0.995** | 0.758 | 0.450 | 0.231 | 0.171 |
| | | LorMC (pairwise loss) | 0.464 | 0.283 | 0.09 | 0.062 | 0.042 |
| | | PRIME (pairwise loss) | 0.377 | 0.246 | 0.028 | 0.021 | 0.015 |
| | | PRIME | 0.266 | 0.125 | 0.021 | 0.017 | 0.005 |
| | 5 | AWTR (Proposed) | 0.548 | 0.384 | **0.264** | 0.219 | **0.207** |
| | | LorSLIM | 0.251 | 0.148 | 0.121 | 0.116 | 0.085 |
| | | SLIM | 0.964 | 0.465 | 0.080 | 0.061 | 0.023 |
| | | NCF | 0.968 | 0.543 | 0.233 | **0.204** | 0.156 |
| | | DeepFM | **0.981** | **0.556** | 0.218 | 0.152 | 0.119 |
| | | LorMC (pairwise loss) | 0.269 | 0.144 | 0.121 | 0.036 | 0.021 |
| | | PRIME (pairwise loss) | 0.283 | 0.138 | 0.051 | 0.021 | 0.014 |
| | | PRIME | 0.22 | 0.116 | 0.024 | 0.019 | 0.007 |
| | 10 | AWTR (Proposed) | 0.320 | 0.233 | 0.167 | 0.156 | **0.147** |
| | | LorSLIM | 0.284 | 0.134 | 0.102 | 0.094 | 0.072 |
| | | SLIM | 0.861 | 0.286 | 0.056 | 0.030 | 0.022 |
| | | NCF | 0.885 | 0.310 | **0.194** | **0.161** | 0.120 |
| | | DeepFM | **0.897** | **0.325** | 0.181 | 0.138 | 0.081 |
| | | LorMC (pairwise loss) | 0.173 | 0.103 | 0.035 | 0.027 | 0.019 |
| | | PRIME (pairwise loss) | 0.173 | 0.106 | 0.029 | 0.021 | 0.015 |
| | | PRIME | 0.116 | 0.082 | 0.027 | 0.018 | 0.011 |
| | 1 | AWTR (Proposed) | 0.811 | 0.640 | **0.337** | **0.192** | **0.149** |
| | | LorSLIM | 0.513 | 0.405 | 0.221 | 0.123 | 0.114 |
| | | SLIM | 0.871 | **0.746** | 0.294 | 0.094 | 0.022 |
| | | NCF | 0.878 | 0.741 | 0.326 | 0.143 | 0.101 |
| | | DeepFM | **0.894** | 0.732 | 0.280 | 0.115 | 0.072 |
| | | LorMC (pairwise loss) | 0.560 | 0.303 | 0.131 | 0.107 | 0.052 |
| | | PRIME (pairwise loss) | 0.494 | 0.230 | 0.027 | 0.019 | 0.011 |
| | | PRIME | 0.306 | 0.141 | 0.022 | 0.017 | 0.009 |
| | 2 | AWTR (Proposed) | 0.694 | 0.611 | **0.323** | **0.184** | **0.136** |
| | | LorSLIM | 0.431 | 0.388 | 0.216 | 0.119 | 0.079 |
| | | SLIM | 0.867 | 0.722 | 0.112 | 0.058 | 0.026 |
| | | NCF | 0.872 | **0.734** | 0.311 | 0.124 | 0.083 |
| | | DeepFM | **0.887** | 0.717 | 0.218 | 0.106 | 0.069 |
| 20 | | LorMC (pairwise loss) | 0.437 | 0.253 | 0.086 | 0.053 | 0.022 |
| | | PRIME ( pairwise loss) | 0.372 | 0.226 | 0.025 | 0.017 | 0.010 |
| | | PRIME | 0.244 | 0.122 | 0.019 | 0.014 | 0.004 |
| | 5 | AWTR (Proposed) | 0.531 | **0.381** | **0.258** | **0.215** | **0.201** |
| | | LorSLIM | 0.243 | 0.136 | 0.112 | 0.108 | 0.072 |
| | | SLIM | 0.851 | 0.379 | 0.103 | 0.047 | 0.022 |
| | | NCF | **0.865** | 0.362 | 0.215 | 0.113 | 0.075 |
| | | DeepFM | 0.853 | 0.341 | 0.176 | 0.101 | 0.061 |
| | | LorMC (pairwise loss) | 0.255 | 0.126 | 0.083 | 0.041 | 0.018 |
| | | PRIME ( pairwise loss) | 0.264 | 0.131 | 0.024 | 0.013 | 0.011 |
| | | PRIME | 0.211 | 0.108 | 0.017 | 0.009 | 0.006 |
| | 10 | AWTR (Proposed) | 0.317 | 0.224 | 0.157 | **0.149** | **0.141** |
| | | LorSLIM | 0.225 | 0.121 | 0.098 | 0.088 | 0.067 |
| | | SLIM | 0.846 | 0.270 | 0.052 | 0.028 | 0.018 |
| | | NCF | 0.853 | 0.286 | **0.128** | 0.102 | 0.064 |
| | | DeepFM | **0.861** | **0.310** | 0.105 | 0.083 | 0.042 |
| | | LorMC (pairwise loss) | 0.165 | 0.095 | 0.027 | 0.021 | 0.009 |
| | | PRIME ( pairwise loss) | 0.167 | 0.101 | 0.021 | 0.014 | 0.010 |
| | | PRIME | 0.112 | 0.075 | 0.022 | 0.011 | 0.006 |

## A.6 Correlation structure settings

Figure A.1 and Figure A.2 represent correlation matrices of patients' and donors' covariates for low (i.e., $\rho = 0$), moderate (i.e., $\rho = 0.5$), and high (i.e., $\rho = 0.8$) correlation structures, respectively. Figure A.3 and Figure A.4 also represent correlation matrices of patients' and donors' covariates for four new block structure correlation scenarios. The black dashed borders in these two figures represent the block of covariates which are highly correlated and not correlated with others.

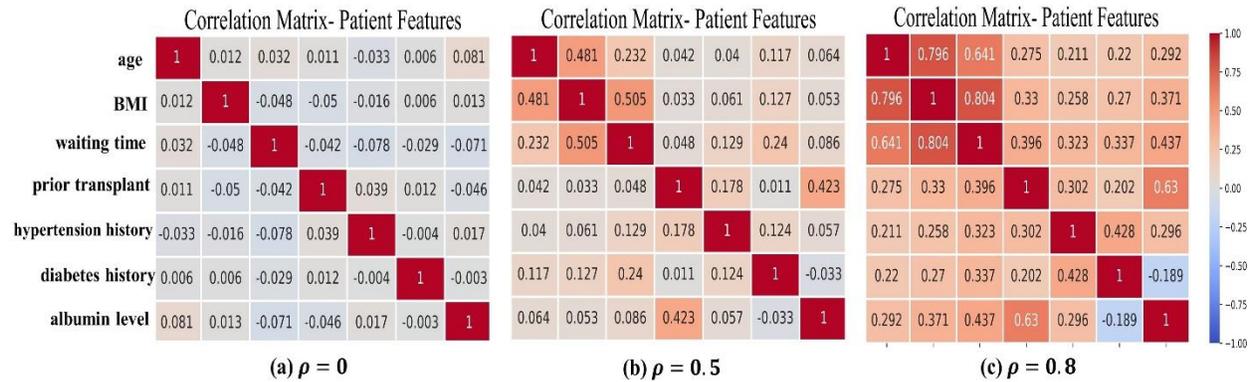

Figure A.1: Correlation matrices of patients' covariates for low, moderate, and high correlation scenarios.

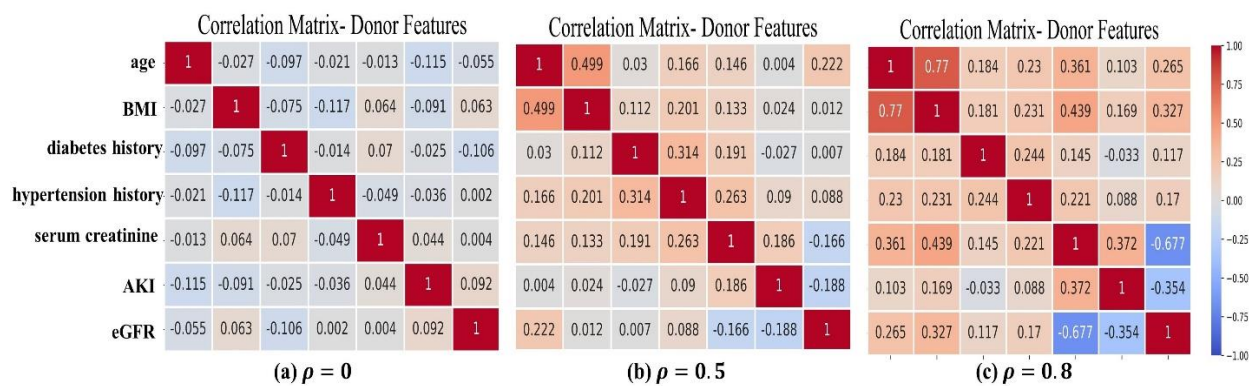

Figure A.2: Correlation matrices of donors' covariates for low, moderate, and high correlation scenarios.

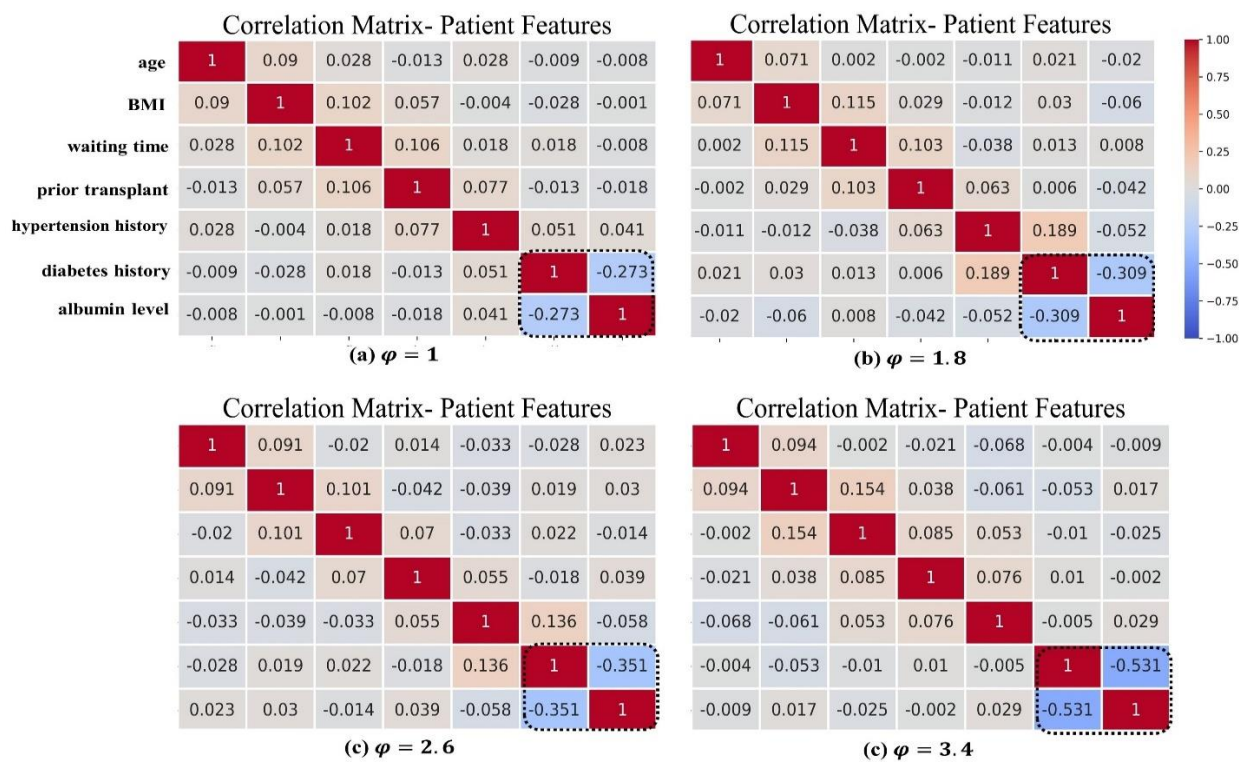

Figure A.3: Correlation matrices of patients' covariates for block structure correlation scenarios.

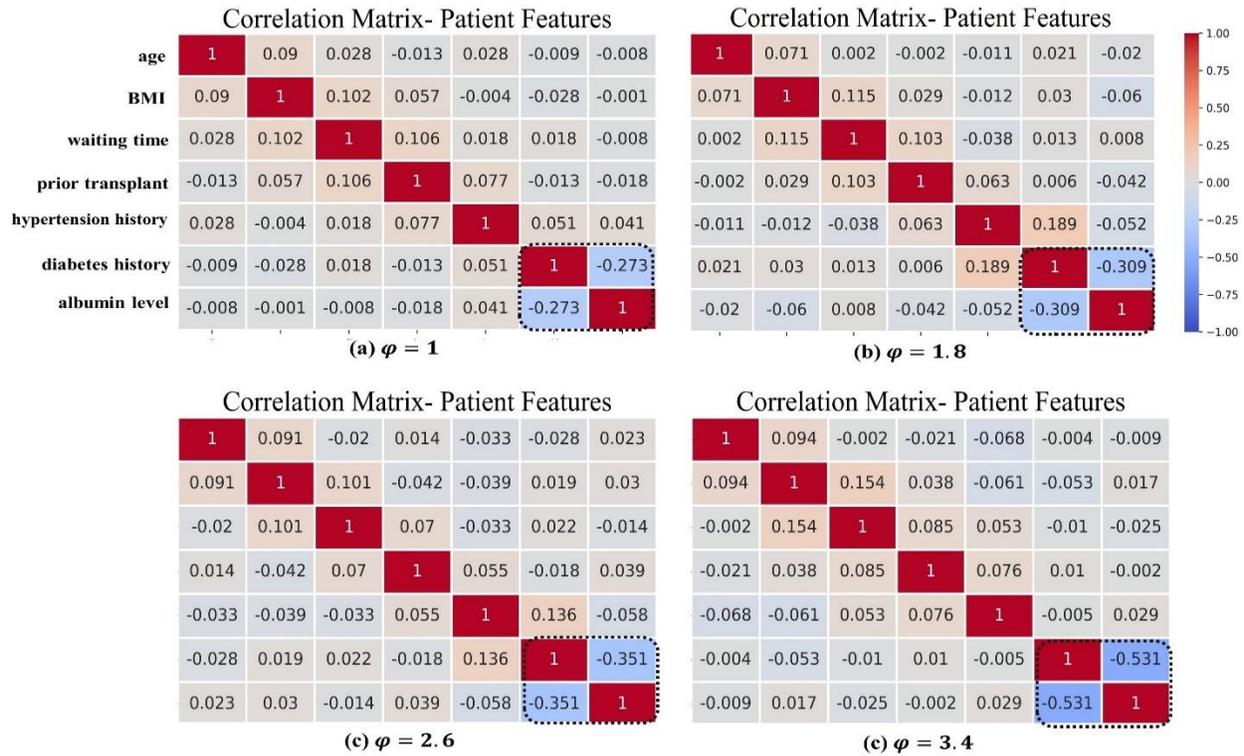

**Figure A.4: Correlation matrices of donors' covariates for block structure correlation scenarios.**